%% file: main.tex
\documentclass{article}

\usepackage{microtype}
\usepackage{graphicx}
\usepackage{subcaption}
\usepackage{booktabs} 

\usepackage{hyperref}

\usepackage[preprint]{icml2026}

\usepackage{amsmath}
\usepackage{amssymb}
\usepackage{mathtools}
\usepackage{amsthm}
\usepackage{bbm}

\usepackage{multirow}
\usepackage[utf8]{inputenc} 
\usepackage[T1]{fontenc}    
\usepackage{hyperref}       
\usepackage{url}            
\usepackage{booktabs}       
\usepackage{amsfonts}       
\usepackage{nicefrac}       
\usepackage{microtype}      
\usepackage{xcolor}         
\usepackage[acronym]{glossaries}
\usepackage{graphicx}
\usepackage{inconsolata}
\usepackage{xspace}
\usepackage{bm}
\usepackage{tikz}
\usepackage{pgfplots}
\pgfplotsset{compat=1.18} 



\usepackage{algorithm}
\usepackage[noend]{algpseudocode}

\renewcommand{\Comment}[1]{\hfill$\triangleright$\ #1}

\usepackage{multirow}
\usepackage{pifont}
\usepackage{caption}

\glsdisablehyper

\newacronym{ar}{AR}{associative recall}
\newacronym{flop}{FLOP}{floating-point operation}
\newacronym{gpt}{GPT}{generative pretrained transformers}
\newacronym{cbs}{CBS}{critical batch size}
\newacronym{twu}{TWU}{token-weighted update}
\newacronym{lm}{LM}{language model}
\newacronym{ih}{IH}{induction head}
\newacronym{la}{LA}{logit attribution}
\newacronym{llm}{LLM}{large language model}
\newacronym{lhs}{LHS}{left hand side}
\newacronym{rhs}{RHS}{right hand side}
\newacronym{lstm}{LSTM}{long short term memory}
\newacronym{mi}{MI}{mechanistic interpretability}
\newacronym{mlp}{MLP}{multi-layer perceptron}
\newacronym{mcmc}{MCMC}{Markov Chain Monte-Carlo}
\newacronym{pwlf}{PWLF}{piecewise linear function}
\newacronym{clm}{CLM}{causal language modeling}
\newacronym{icl}{ICL}{in-context learning}
\newacronym{iwl}{IWL}{in-weight learning}
\newacronym{ppp}{PPP}{psychometric predictive power}
\newacronym{bert}{BERT}{bidirectional encoder representations from transformers}
\newacronym{ps}{PS}{prefix-matching score}
\newacronym{sas}{SAS}{syntactic attention structure}
\newacronym{snr}{SNR}{signal-to-noise ratio}

\usepackage[capitalize,noabbrev]{cleveref}

\theoremstyle{plain}

\theoremstyle{definition}

\theoremstyle{remark}

\usepackage[textsize=tiny]{todonotes}

\icmltitlerunning{Predicting the Emergence of Induction Heads in Language Model Pretraining}

\begin{document}

\twocolumn[
  \icmltitle{Predicting the Emergence of Induction Heads\\in Language Model Pretraining}



  \icmlsetsymbol{equal}{*}

  \begin{icmlauthorlist}
    \icmlauthor{Tatsuya Aoyama}{yyy}
    \icmlauthor{Ethan Gotlieb Wilcox}{yyy}
    \icmlauthor{Nathan Schneider}{yyy}
  \end{icmlauthorlist}

  \icmlaffiliation{yyy}{Department of Linguistics, Georgetown University}

  \icmlcorrespondingauthor{Tatsuya Aoyama}{ta571@georgetown.edu}

  \icmlkeywords{Machine Learning, ICML}

  \vskip 0.3in
]



\printAffiliationsAndNotice{}  

\begin{abstract}
Specialized attention heads dubbed \textit{induction heads} (IHs) have been argued to underlie the remarkable in-context learning capabilities of modern language models; yet, a precise characterization of their emergence, especially in the context of language modeling, remains wanting. In this study, we investigate the relationship between statistical properties of the training data and IH formation in both natural and synthetic training data settings. We show that: (1)~a simple equation combining batch size and context size predicts the point at which IHs form and that this emergence point is agnostic to model size; (2)~surface bigram repetition frequency and reliability strongly affect the formation of IHs, and we find an effective decision boundary in terms of these two values; (3) local dependency with high bigram repetition frequency and reliability is sufficient for IH formation, but categoriality and the shape of the marginal distribution appear to modulate IH formation near the decision boundary.
\end{abstract}

\section{Introduction}\label{ch5:sec:intro}

The practical utility of modern \glspl{llm} depends heavily on their ability to perform \gls{icl}, broadly construed as performing a task based on the input provided at inference time. Various accounts have been provided to explain the internal workings of this capability, and among them are studies that find certain attention heads affecting \glspl{lm}' \gls{icl} capabilities \cite{olsson2022context, edelman2024the, reddy2024the, yin2025function}. \citet{olsson2022context} maintain that specialized heads that emerge abruptly during pretraining, dubbed \glspl{ih}, are primarily responsible for the emergent \gls{icl} capabilities \glspl{lm} exhibit. \citet{edelman2024the} mathematically show \gls{icl} is equivalent to a type of Bayesian inference, and that \glspl{ih} are indispensable in making this inference. The emergence of such heads during pretraining is sometimes referred to as phase change or phase transition \cite{chen2024sudden, aoyama-wilcox-2025-language}. The broader phenomenon of phase transition is not limited to the text modality, and has been observed in vision models \cite{okawa2023compositional, park2024emergence}.

It has been shown that \glspl{ih} form in \glspl{lm} very early in pretraining, regardless of the model size, as long as the model has multiple heads and at least 2 layers \cite{olsson2022context}. While the behavior and mechanism of \glspl{ih} are relatively well-understood \cite{elhage2021mathematical, olsson2022context, edelman2024the}, it is not clear \textit{when} and \textit{why} such heads form naturally when trained on language data. In other words, we ask: when exactly do \glspl{ih} emerge in \glspl{lm}, and what properties of natural language give rise to \gls{ih} formation, potentially by incentivizing the copying behavior in transformer \glspl{lm}? Given that the emergence of \glspl{ih} has downstream behavioral consequences, precisely characterizing their emergence point has practical implications for \gls{lm} pretraining, such as dataset selection and training efficiency.

In this study, through a series of experiments using both natural and synthetic data, we draw the following conclusions: (1)~a simple equation combining batch size and context size can predict the point at which \glspl{ih} form and that this emergence point is agnostic to model size (\Cref{ch5:sec:predict}); (2)~the surface bigram repetition frequency and reliability strongly affect the formation of \glspl{ih}, and a decision boundary can be described in terms of these two values (\Cref{ch5:sec:exp2}); (3)~local dependency with high bigram repetition frequency and reliability is sufficient for IH formation, but categoriality and the shape of the marginal distribution appear to modulate \gls{ih} formation near the decision boundary (\Cref{ch5:sec:exp3}). Finally, we also show that the emergence points relate non-monotonically to final validation loss, suggesting their relevance to broader training dynamics (\Cref{ch5:sec:dynamics}).\footnote{Code available at \url{https://github.com/t-aoyam/predict-ih/}.}\footnote{We will use emergence, phase change, and phase transition interchangeably in this paper.}

\section{Relevant Work}\label{ch5:sec:relevant}

Factors that affect phase transition, particularly the formation of \glspl{ih}, are not fully understood. \glspl{ih} are often associated with \glspl{lm}' \gls{icl} capabilities.\footnote{\citet{olsson2022context} claim that \glspl{ih} are primarily responsible for \gls{icl}, whereas \citet{yin2025function} find that a different set of heads, dubbed \textit{function vector heads}, are performing few-shot-learning-style \gls{icl}.} \citet{chan2022data} study the data properties that lead to different learning outcomes in few-shot image classification, where few-shot examples are image-label pairs. They find that (1) burstiness (similar things appearing in clusters), (2) within-class variation, and (3) dynamic class membership all promote \gls{icl} and demote \gls{iwl}. Interestingly, making the marginal distribution of the labels Zipfian was the only variable that led to high \gls{icl} and \gls{iwl} simultaneously.

\citet{edelman2024the} propose a synthetic task dubbed \gls{icl} Markov Chain (\gls{icl}-MC) using a Markov Process involving 2-8 symbols. Taking a Bayesian approach to \gls{icl} \cite{xie2022an}, \citet{edelman2024the} randomly initialize the transition matrix at the beginning of each epoch, thereby making it impossible to learn the underlying distribution of the symbols. They find that the models sequentially go through phases where their predictions most closely match the uniform distribution, then in-context unigram counts, and lastly in-context bigram counts. They show that by forming \glspl{ih} that attend to all of the bigram continuations of the current token in the preceding context, models achieve the Bayes-optimal bigram solution.

\Gls{icl} is often studied as few-shot learning (e.g., \citealp{chan2022data,singh2024what}) with input-label (e.g., image-label) pairs, or with a synthetic setting that necessitates \gls{icl} as opposed to \gls{iwl} (e.g., \citealp{edelman2024the,park2025competition}). \citet{park2025competition} propose a novel synthetic sequence modeling task using a mixture of Markov chains; however, in \gls{lm} pretraining, it is unlikely that each batch comes from a completely different distribution of tokens. As such, it is yet to be clear why \glspl{lm}, which are trained on natural language through next token prediction, form \glspl{ih} (and \gls{icl} capabilities), and what properties of natural language promote such learning dynamics. Furthermore, the point at which \gls{ih} formation occurs is not fully understood. Some report a narrow range (1B--3B pretraining tokens; e.g., \citealp{olsson2022context}) while others find a much wider range (64M--2B pretraining tokens; \citealp{aoyama-wilcox-2025-language}). \citet{zucchet2025the} study the emergence point; however, they focus on a general framework called sparse attention using a synthetic associative recall task. We focus specifically on \glspl{ih} in the naturalistic language modeling task.

To fill these gaps, in this study, we aim to find (1) the point at which \glspl{ih} emerge, and (2) precise data properties that promote the formation of \glspl{ih}.

\section{Methods}\label{ch5:sec:method}

\subsection{Metrics} \glspl{ih} are defined by the copying behavior. If the model has seen an \ab{A}{B} sequence \textit{in-context} and the current token is \tok{A}, then an \gls{ih} is a head that promotes the prediction of \tok{B} as the next token, completing the \abab{A}{B} sequence.

\textbf{Prefix-matching Score.} Following \citet{olsson2022context}, we quantify this behavior using \gls{ps}. Given a random sequence of tokens $\mathbf{x}$ repeated twice, \gls{ps} of a head $h$ at layer $l$ is its average attention from the source token $x_i$ to the next token of its previous occurrence:
\begin{equation}
\frac{1}{|\mathbf{x}|}\sum_{i=|\mathbf{x}|+1}^{2|\mathbf{x}|}\mathrm{\alpha}^{(h,l)}(x_i, x_{i-(|\mathbf{x}|-1)})
\end{equation}
As for the size of the sequence $\mathbf{x}$, \texttt{TransformerLens} library \cite{nanda2022transformerlens} adopts $|\mathbf{x}|=50$, which we also do; however, when analyzing models with smaller context sizes, we adjust $|\mathbf{x}|$ accordingly: $|\mathbf{x}|=\min (\frac{\text{|context|}}{2}, 50)$.

\textbf{Logit Attribution.} Similarly, \citet{olsson2022context} define a metric called \gls{la}. As opposed to \gls{ps}, which measures a given head's attention to the token of interest, \gls{la} measures the actual contribution of a given head's output to the model's final logit of the token of interest via the residual stream. The raw logit contribution $C\in \mathbb{R}^{|\mathcal{V}|}$ of an output $x^{h, l}$ from a given head $h$ at layer $l$ is computed by passing it through the linear layer of the attention block $W_O$ and the final unembedding layer $U$:
\begin{equation}
    (x^{h,l}W_O)U
\end{equation}

Following \citet{olsson2022context}, we then center and normalize the positive logit contribution to all tokens in the test sequence $\mathbf{x}$ and compute the ratio of the logit contribution to the target token to the rest of the tokens in the sample. Note that since all models used in this study (GPT2, Pythia) employ pre-layer normalization, there exists a direct path from a given head's output to the final logit calculation. \Gls{la} captures each head's contribution through this path. However, it is important to note that the rest of the contribution (e.g., through subsequent layers) is not captured in \gls{la}.

\textbf{Associative Recall.} Lastly, we also measure how often a model predicts \tok{B} given \aba{A}{B}, a metric referred to as \gls{ar}. It can be measured by accuracy:
\begin{equation}
    \frac{1}{|\mathbf{x}|}\sum_{i=|\mathbf{x}|+1}^{2|\mathbf{x}|}\mathbbm{1}\{f(\mathbf{x}_{<i})=x_i\}
\end{equation}
and by the mean rank of the target token \tok{B}:
\begin{equation}
    \frac{1}{|\mathbf{x}|}\sum_{i=|\mathbf{x}|+1}^{2|\mathbf{x}|}\text{rank}(x_i;f(\mathbf{x_{<i}}))
\end{equation}
As opposed to the previous two metrics, which are head-level, these metrics are model-level. As reported later, the abrupt improvement in \gls{ar} always follows that in \gls{ps} and \gls{la}, in line with the findings from \citet{reddy2024the}. Also, given a high correlation between \gls{ps} and \gls{la}, we report the analyses of \gls{ps} in the main body of this paper.

\subsection{Models and Checkpoints} At least 2 attention layers have been shown to be necessary for the model to perform the induction task \cite{olsson2022context, reddy2024the, ekbote2025what} unless the model's hidden dimension is increased exponentially \cite{sanford2024one}. In fact, models with 2 layers have recently been shown to be sufficient to approximate any-order Markov chain \cite{ekbote2025what}. Given these findings, we use a 50M-parameter GPT2 \cite{radford2019language} with 2 layers and 8 attention heads per layer with the hidden dimension of 768 for all experiments. For the first experiment (\Cref{ch5:sec:exp1}), to study the effect of model size, we also train larger models with 125M parameters and 350M parameters and include models up to 7B parameters for inference. See \Cref{tab-hyperparams} for the details on model architecture and training setup.

For experiments with natural language (\Cref{ch5:sec:exp1}), we adopt a pretrained GPT2 tokenizer with a vocabulary size of 50,257, unless otherwise specified. All models analyzed in this study were trained from scratch for 1B pretraining tokens. We save intermediate checkpoints at 250K and 500K tokens, [1M, 10M) tokens at 1M increments, [10M, 100M) tokens at 10M increments, and [100M, 1B] tokens at 100M increments, resulting in 30 checkpoints per model. We additionally use pretrained Pythia models \cite{biderman2023pythiasuiteanalyzinglarge} for follow-up analyses. Note that Pythia models were deemed the only available model family with early enough checkpoints available to study the emergence of \glspl{ih}, which is known to happen very early in pretraining (64M-3B pretraining tokens; \citealp{olsson2022context, aoyama-wilcox-2025-language}). For example, OLMo models provide checkpoints for every 1000 steps of pretraining \cite{groeneveld2024olmoacceleratingsciencelanguage}, meaning that the earliest model checkpoint is already trained on 4B tokens, likely after the phase transition of interest.

For experiments with synthetic language (\Cref{ch5:sec:exp2,ch5:sec:exp3}), we adopt a vocabulary size of 10,000. We do not use a tokenizer as we only work with token IDs. Note that the change in vocabulary size will likely affect the emergence points \cite{singh2024what,zucchet2025the}; however, this should not change the results of our analyses in these experiments, as we only focus on \textit{whether} \glspl{ih} emerge or not, and not on \textit{when} they do, in \Cref{ch5:sec:exp2} and \Cref{ch5:sec:exp3}.

\subsection{Data} For natural texts (\Cref{ch5:sec:exp1}), unless otherwise specified, we use the English subcorpus from the Common Crawl Corpus (\texttt{CC100}; \citealp{conneau-etal-2020-unsupervised, wenzek-etal-2020-ccnet}). We create a sample of 1B tokens from this corpus, using the pretrained GPT2 tokenizer. As mentioned earlier, all models are trained for 1 epoch on this sample, for a total of 1B tokens. For semi-natural data (\Cref{ch5:sec:exp2}) and synthetic data (\Cref{ch5:sec:exp3}), we use a token-to-token transition matrix, and the details are provided in each section.
\begin{figure}[t]
\centering
\input{figures/ps_la_acc_mr.pgf}
    \caption[Developmental trajectories of PS, LA, AR accuracy, and AR mean rank.]{Developmental trajectories of PS, LA, AR accuracy, and AR mean rank. The first three metrics are plotted in the scale on the left $y$-axis, and the last metric in the scale on the right $y$-axis.}
    \label{ch5:fig:ps_la_acc_mr}
\end{figure}
\begin{figure*}[ht] 
    \centering
    \input{figures/knot_slope_with_curves.pgf}
    \caption[Developmental trajectories of \gls{ps} of \glspl{lm} with various batch sizes (left), context sizes (center), and repetitions (right) over the course of 1B tokens of pretraining.]{Developmental trajectories of \gls{ps} of \glspl{lm} with various batch sizes (left), context sizes (center), and repetitions (right) over the course of 1B tokens of pretraining, plotted against the number of updates (a) and the number of tokens (b). BS, CS, \%R stand for batch size, context size, and the proportion of chunks with natural bigram repetitions, respectively. The bottom figure plots the identified ``emergence point'' on the $x$-axis and the slope of the corresponding segment of the curve on the $y$-axis. Configurations that did not lead to the emergence of \glspl{ih} were excluded from the bottom plot (i.e. context size 4, 8, 16, and \%R 30).}
    \label{ch5:fig:batch_context_rep_ps_steps}
\end{figure*}
\section{Metric Selection}\label{ch5:sec:metric}
Before diving into the main experiments, as briefly mentioned earlier, we show how \gls{ps}, \gls{la}, and \gls{ar} develop during pretraining. \Cref{ch5:fig:ps_la_acc_mr} shows a sample model (GPT2 125M) trained with batch size of 16 and context size of 128. Clearly, all four metrics go through an abrupt change at around the same time. Notably, \gls{ar} accuracy seems to increase slightly later compared to the other three metrics. It is important to recall that the best \gls{ps} and best \gls{la} reflect head-level changes in the strategy a model is employing, and that \gls{ar} (mean rank) measures the model's change in the direction of predicting the target token \tok{B}, even when it is not the most probable token. On the other hand, \gls{ar} (accuracy) only improves when the target token \tok{B} is the most probable token; in other words, the model's improvement from ranking the token \tok{B} as the least probable (rank 50,257) to second most probable (rank 2) is not captured in this metric. Given these observations, for simplicity and readability, we only report \gls{ps} in the main body of this paper, while replicating similar analyses with other metrics in \cref{appendix:results-other-metrics}.

\section{Experiment 1: Natural Data}\label{ch5:sec:exp1}

\subsection{Methods}

As briefly introduced earlier, \citet{aoyama-wilcox-2025-language} find that training an \gls{lm} with different batch sizes results in different phase transition points. In this experiment, we change context size and batch size to investigate their effect on the formation of \glspl{ih}. Specifically, we experiment with log-spaced batch and context sizes ranging from 4 to 512 and 4 to 2048, respectively. Since grid search is expensive, we fix the batch size at 16 while changing the context size, and fix the context size at 1024 while changing the batch size. We additionally test the effect of bigram repetitions alone by selecting subsets of the pretraining data.

\subsection{Results}
To describe the effect of each of the 3 variables, we identify 2 types of effect: \textit{shifting} and \textit{slanting}. \textit{Shifting} is the change in the point at which \glspl{ih} emerge, which we identify by fitting a \gls{pwlf} to the \gls{ps} curve and taking the first knot as the emergence point. \textit{Slanting} is the change in the slope of the fitted \gls{pwlf} after the emergence (\Cref{appendix:fitting-procedure-emergence}). \Cref{ch5:fig:batch_context_rep_ps_steps} (bottom) plots the emergence points on the $x$-axis and the slope on the $y$-axis; hence, a movement along the $x$-axis corresponds to the \textit{shifting} effect, and movement along the $y$-axis corresponds to the \textit{slanting} effect.

\textbf{Batch Size.} In \Cref{ch5:fig:batch_context_rep_ps_steps} (top left), we find (1) the larger the batch size, the lower the eventual \gls{ps}; in other words, a larger batch size leads to weaker \glspl{ih} at the end of the pretraining, and (2) the smaller the batch size, the later the ``spike'' in \gls{ps}; in other words, training an \gls{lm} with a smaller batch size results in later emergence of \glspl{ih}, as measured
by the number of updates. The bottom plot confirms this shifting effect: decreasing the batch size has little and mixed effects on the slope (i.e., no slanting; $\rho=0.29,p=0.49$), but a monotonically positive effect on the emergence point ($\rho=-1.0, p<0.001$).

\textbf{Context Size.} In \Cref{ch5:fig:batch_context_rep_ps_steps} (center), we find that (1) the smaller the context size, the later the emergence (shifting), and (2) the smaller the context size, the flatter the slope (slanting), and the extreme case (context size $\le 16$) is a flat line, or the complete suppression of \glspl{ih} (see \Cref{appendix:mcmc} on how we determine ``random'' attention). Both of these effects are confirmed in the bottom plot, where the emergence point is monotonically increasing ($\rho=-1.0, p<0.001$) and the slope is monotonically decreasing ($\rho=1.0, p<0.001$) as we reduce the context size. For the shifting effect in (1), since this effect was observed both for batch and context size, we suspect that this could be attributed to the number of tokens the model is exposed to at each update. For the slanting effect in (2), we suspect that, in natural texts, a larger context size will naturally contain more occurrences of \abab{A}{B} patterns, which may have a threshold below which \glspl{ih} will not form.

Note that we observe an inverse shifting effect when plotting against the number of pretraining \textit{tokens} in \Cref{ch5:fig:batch_context_rep_ps_toks} in \Cref{appendix:tokens_not_steps}, thereby ruling out the possibility that the observed shifting effect is an artifact of each point on the $x$-axis representing a different number of pretraining \textit{tokens}.

\textbf{Repetition.} The number of bigram repetitions increases as context size grows (see \Cref{ch5:fig:nat_dist} in \Cref{appendix:nat-rep}), which we hypothesize to cause the slanting effect of the context size. To tease apart these two phenomena, we manipulate the occurrence rate of bigram repetitions within each chunk, while controlling for the batch size and context size. See \Cref{appendix:nat-rep} for more details on how we manipulate the repetition rate while controlling for the context size. In \Cref{ch5:fig:batch_context_rep_ps_steps} (right), (1) the higher the proportion of chunks with bigram repetitions, the higher the best \gls{ps} a given model achieves, and (2) we only observe the slanting effect and no shifting effect. This is again confirmed in the bottom plot, where the effect on the emergence point is insignificant ($\rho=-0.64,p=0.12$), but the effect on the slope is significant ($\rho=0.93, p=0.002$). This confirms the observation earlier that the shifting is due to the number of tokens an \gls{lm} sees per update, and slanting is due to the rate with which an \gls{lm} encounters repeated bigrams.

\section{Predictability and Robustness}
\subsection{Predictive Law}\label{ch5:sec:predict}
We have seen that context size and batch size affect the phase transition point. Here, we ask: can we predict the phase transition point only using the training configuration as variables (i.e., before we train the model)? We start with a full regression model that predicts the number of updates at which \glspl{ih} emerge based on batch size, context size, and model size. However, as the model size turns out to be the only \textit{non}-significant predictor (see \Cref{appendix:fitting-procedure} for details), we proceed with a regression model without model size:
\begin{equation}\label{ch5:eq:general}
    \begin{aligned}
    U_{\textsc{pt}}= e^\alpha B^\beta C^\gamma,
\end{aligned}
\end{equation}
where $B$ and $C$ represent \textbf{B}atch size, \textbf{C}ontext size, respectively, and $\beta$, and $\gamma$  are their corresponding coefficients. $e^\alpha$ serves as an intercept, as we will see below.
Following \citet{kaplan2020scalinglawsneurallanguage}, we estimate the parameters $\alpha$, $\beta$, and $\gamma$ using a simple ordinary least squares regression in log space and obtain $\alpha=13.26$, $\beta=-0.37$, and $\gamma=-0.62$ (see \Cref{tab:nomodelsize_regression} for all test statistics). For notational simplicity, let $e^\alpha=T$. Plugging these parameters back into \Cref{ch5:eq:general}, we obtain:
\begin{equation}
\begin{aligned}[b]\label{ch5:eq:constant}
    U_{\textsc{pt}}&= \frac{T}{B^{0.37} C^{0.62}}\\
    T&= U_{\textsc{pt}}B^{0.37} C^{0.62}
\end{aligned}
\end{equation}
The key intuition behind \Cref{ch5:eq:constant} is that the model-agnostic constant $T$ is a function of the \textit{quantity} of training, or the number of updates $U_{\textsc{pt}}$, at which phase transition occurs. At the same time, the \textit{quality} of each update matters, and it correlates with the number of tokens the model sees at each update, which is a function of context and batch sizes $B^{0.37} C^{0.62}$. We call the generalized form of the \gls{rhs} of this equation, $UB^{0.37} C^{0.62}$, the number of \glspl{twu}, given that it is the number of updates scaled by batch size and context size to incorporate the number of tokens seen at each update. \Cref{ch5:eq:constant} suggests that the number of \glspl{twu} at which phase transition occurs can be expressed as a constant $T$ across model and training configurations. To further verify that this simple law indeed predicts the phase transition point of \glspl{lm} trained with various training configurations, we can reformulate \Cref{ch5:eq:constant} to predict the number of tokens $N$. Given that $N=UBC$ by definition, we get:
\begin{equation}\label{ch5:eq:pred}
    N_{\textsc{pt}}=TB^{0.63} C^{0.38}
\end{equation}
\begin{figure}[t]
\centering
\input{figures/twu_pred_vs_obs.pgf}
    \caption[Predicted ($x$-axis) and observed ($y$-axis) numbers of pretraining tokens at which phase transition occurs.]{Predicted ($x$-axis) and observed ($y$-axis) numbers of pretraining tokens at which phase transition occurs. A strong correlation of $r=.986$ ($p<.001)$ is found.}
    \label{ch5:fig:twu_pred_vs_obs}
\end{figure}
The \gls{lhs} of \Cref{ch5:eq:pred} is the \textit{observed} number of pretraining tokens at which phase transition occurs, and the \gls{rhs} is the \textit{predicted} point based on context size $C$ and batch size $B$, as well as the empirically found constant $T=e^{13.26}$. Now we can predict the number of tokens $N$ at which phase transition occurs, based on a constant $T$ and training configurations $C$ and $B$. In \Cref{ch5:fig:twu_pred_vs_obs}, $x$-axis and $y$-axis correspond to the \gls{lhs} and \gls{rhs}, or the predicted and observed number of pretraining tokens at which phase transition occurs, respectively. We find a strong correlation of $r=.986$ ($p<.001)$. This law holds for more than 3 orders of magnitude in model size (50M--7B), and 5 orders of magnitude in the number of tokens per update (500--2M).
\subsection{Robustness}\label{ch5:sec:robust}
Given that the models presented thus far are based on a single random seed, and that the emergence point estimation through \gls{pwlf} from each run could be noisy, we train with 2 additional random seeds the following 6 representative batch-context configurations: 16-64, 16-256, 16-1024, 16-2048, 128-1024, and 512-1024, which allows us to report the variability across 3 random seeds.
\begin{figure}[t]
  \begin{subfigure}[t]{0.49\linewidth}
    \centering
    \makebox[\linewidth][l]{\textbf{(a)}}\\
    \input{figures/emergence_var.pgf}
    \phantomcaption
    \label{ch5:fig:emergence_var}
  \end{subfigure}
  \begin{subfigure}[t]{0.49\linewidth}
    \centering
    \makebox[\linewidth][l]{\textbf{(b)}}\\
    \input{figures/law_var.pgf}
    \phantomcaption
    \label{ch5:fig:law_var}
  \end{subfigure}
      \caption{\textbf{(a)} \gls{ih} emergence points measured in the number of updates and their 95\% confidence intervals across 3 random seeds. Each configuration on the $x$-axis is a batch-context pair. \textbf{(b)} Predicted ($x$-axis) and observed ($y$-axis) numbers of pretraining tokens at which phase transition occurs. The color and shape of each point represent the random seed, and the gray shaded area represents data points from the same batch-context configuration across 3 different random seeds.
}
\end{figure}
We can see in \Cref{ch5:fig:emergence_var} that each configuration has tight 95\% confidence intervals, suggesting robustness across different random seeds.

Similarly, we fit the predictive law, as described earlier (\Cref{ch5:sec:predict}), on the set of 6 configurations for each of the 3 random seeds. \Cref{ch5:fig:law_var} shows that the predictive law $TB^{\beta+1}C^{\gamma+1}$ ($x$-axis) captures the observed emergence points ($y$-axis) well regardless of the random seed, showing its robustness. We report the full results of this uncertainty estimation in \Cref{appendix:var}.

\section{Experiment 2: Semi-Natural Data Using Natural Bigram Statistics}\label{ch5:sec:exp2}

\subsection{Methods}\label{ch5:exp2:methods} In \Cref{ch5:fig:batch_context_rep_ps_steps} (right), we manipulated the repetition rate at the chunk level; in other words, we only manipulated the proportion of chunks with at least one bigram repetition. This repetitiveness at the chunk level is treated as a discrete variable called \textit{burstiness} in \citet{chan2022data} and extended to a continuous variable in \citet{reddy2024the} for an in-context classification task. \citet{zucchet2025the} also report the effect of repetition in training data; however, this is again at the sequence level and in an associative recall task, as opposed to language modeling task. To more precisely manipulate the repetition rate in the context of language modeling, we define two metrics, \textit{frequency} and \textit{reliability}.

\textit{Frequency} measures the relative frequency at which \aba{A}{B} is observed in a given dataset, expressed as \paba{A}{B}. It is simply the proportion of tokens in a given dataset that complete a \aba{A}{B} sequence as the second occurrence of \tok{A}, where $\tok{A}\ne\tok{B}$.

\textit{Reliability} measures the conditional probability with which \tok{B} is observed given \aba{A}{B}, expressed as \pbgivenaba{A}{B}. It is the proportion of tokens in a given dataset that complete a \abab{A}{B} sequence as the second occurrence of \tok{B}, where $\tok{A}\ne\tok{B}$, divided by the aforementioned \textit{frequency}. See \Cref{appendix:freq_and_rel} for more details on these two metrics, and \Cref{ch5:fig:abab_probas} for how these two metrics change for various context sizes in natural data.
\begin{figure}[t]
    \centering
    \input{figures/exp2.pgf}
    \caption{Best \gls{ps} across all heads at the end of the training for each frequency reliability combination, observed (left) and predicted by the fitted model (right). Scores are represented in colors, with brighter colors representing higher scores.}
    \label{ch5:fig:nat_grid}
\end{figure}
\begin{table*}[t]
    \centering
    \caption{Markov Processes used for pretraining data generation in Experiment 3. Each row represents a matrix that defines the Markov Process. The \textbf{Properties} column lists desired properties the matrix was optimized for, and the \textbf{Statistics} column summarizes the actual statistical properties each matrix had at the end of the optimization process. LD and CAT represents the binary variables local dependency and categoriality, respectively. H measures the entropy of the data, and KL divergence measures the fit between the desired distribution (as shown in the \textbf{Properties} column) and the actual marginal distribution of the generated matrix.}
    \input{tables/dists}
    \label{ch5:tab:dists}
\end{table*}

To create training data that resemble natural data and also satisfy desired values for these two metrics, we first generate a token-to-token transition matrix based on the bigram statistics from CC100. We then sample from this matrix, while imposing the specified frequency-reliability configuration, and train an \gls{lm} for each configuration. See \Cref{appendix:algo} for more details on the sampling procedure. We also note that the two properties are imposed on the second half of the sequence, rather than the entire sequence. We use the context size of 64 in this experiment.

\subsection{Results}\label{ch5:exp2:results} Each cell of \Cref{ch5:fig:nat_grid} represents an \gls{lm} trained in the configuration specified by the $x$ and $y$ axes. We conduct a grid search over a search space defined by all possible combinations of each metric ranging from \{0.1, 0.3, 0.5, 0.7, 0.9\}. We initially found that the formation of \glspl{ih} was insensitive to different values of \pbgivenaba{A}{B} when $\paba{A}{B}\ge 0.1$, and hence conducted an additional grid search over $\paba{A}{B}\in \{0.01, 0.03, 0.05, 0.07, 0.09\}$, as well as $\pbgivenaba{A}{B}=0.2$. With a total of 60 models colored based on the best \gls{ps} in \Cref{ch5:fig:nat_grid}, we can clearly see a decision boundary, where a decrease in either value will result in the failure of \gls{ih} emergence.

Notably, \glspl{lm} studied here seem to show stronger sensitivity to reliability than to frequency. In the bottom half of \Cref{ch5:fig:nat_grid}, [$p_1$, $p_2$] and [$p_2$, $p_1$] do not always show the same result. For example, $\paba{A}{B}=0.1$ always leads to \gls{ih} formation except for $\pbgivenaba{A}{B}\in\{0.1,0.2\}$; however, it never forms when $\pbgivenaba{A}{B}=0.1$.
This is notable, given that [$p_1$, $p_2$] and [$p_2$, $p_1$] have identical numbers of bigram repetitions because $\pabab{A}{B}=\pbgivenaba{A}{B}\paba{A}{B}$.
\subsection{Predictability}
To quantify the effect of frequency and reliability on the emergence of \glspl{ih}, we fit a predictive model $\sigma(k(\alpha\log(P_A)+\beta\log(P_B)-\tau))$, where $P_A$ and $P_B$ are \paba{A}{B} and \pbgivenaba{A}{B}, respectively. We obtain $k=4.299,\alpha=0.472,\beta=1.251,\tau=-2.322$, and the fitted model predicts the observed values well as shown in \Cref{ch5:fig:nat_grid} (right), with an MSE of 0.019. We can see that the best \gls{ps} is more than twice as sensitive to reliability as it is to frequency, matching our earlier observation. Lastly, plugging in the fitted values and solving for $P_B$, we obtain a functional form: $P_B = 0.156\times {P_A}^{-0.378}$.

\section{Experiment 3: Synthetic Data}\label{ch5:sec:exp3}

The previous experiment relied on a Markov Process obtained from a naturally occurring text (i.e., \texttt{CC100}). The main goal of this last experiment is to describe the properties of the underlying Markov Process necessary and/or sufficient for \gls{ih} formation.

\subsection{Methods} For simplicity, and to allow for more precise control over the data properties, we limit our scope to the second order Markov Process as the underlying generative process, which can be expressed as a token-to-token transition matrix $T\in\mathbb{R}^{|\mathcal{V}|\times|\mathcal{V}|}$. Once the desired properties (see below) are specified, we optimize the matrix using the Adam optimizer (see \Cref{ch5:appendix:optimization} for details). We consider three properties that we hypothesize to affect the formation of \glspl{ih}: (1) local dependency, (2) categoriality, and (3) the shape of the marginal distribution. For (1) local dependency ($\pm\text{D}$), it is construed as: $+\text{D}$ iff $P(w_{t+1}|w_t)\space\ne\space P(w_{t+1})$. In other words, unless the random variable $W$ is i.i.d. at each position $t$, we consider the distribution $+$D. This simply means that a distribution is $+$D if a word affects the next word.

For (2) categoriality ($\pm\text{C}$), as \glspl{ih} have been shown to copy abstract patterns, such as semantic categories (e.g., color-object sequences; \citealp{olsson2022context}), we suspect that the presence of categories promotes the formation of \glspl{ih}. To make this property compatible with the optimization process, we define categoriality by inter-group and within-group similarity scores. We define the presence ($+\text{C}$) and absence ($-\text{C}$) of categoriality as having within-category similarities of 0.4 and 0.1, respectively. Between-category similarity was always set to 0.1. We first assign $\frac{|\mathcal{V}|}{|\mathcal{C}|}$ tokens into each category $c\in \mathcal{C}$, thereby creating $|\mathcal{C}|$ groups with disjoint members. We then define inter-group similarity as the average cosine similarity between words (i.e., each of the $|\mathcal{V}|$ rows of the transition matrix $\mathcal{T} \in \mathbb{R}^{|\mathcal{V}|\times|\mathcal{V}|}$) from a given category $c$ and words from a different category $c'$:
\begin{equation}
    \frac{1}{N}\sum_{c, c'\in \mathcal{C}, \space c\ne c'}\sum_{w\in c}\sum_{w'\in c'}{sim(w,w')}
\end{equation}
where $N$ is the number of such word pairs. Within-category similarity is likewise defined as the average similarity between all pairs of words that belong to the same category. If a distribution has a high within-category similarity but a lower across-category similarity, it means that categories exist in this distribution.

\begin{figure}[t]
    \centering
    \input{figures/exp3_ps.pgf}
    \caption{\gls{ih} formation, as measured by PS, in each of the pretraining data generated by the Markov Processes. Each column represents a frequency-reliability configuration.}
    \label{ch5:fig:exp3-ps}
\end{figure}

Lastly, for (3) marginal token distribution shape, we consider 3 distribution shapes that are increasingly less uniform: Uniform, Gaussian, and Zipfian distributions. This is because natural language is uniquely characterized by a Zipfian distribution, an inverse power law that expresses the frequency of a given word as inversely correlated with its rank, and that it has been shown to affect the emergence of \gls{icl} capabilities (e.g., \citealp{chan2022data}).

Taken together, we have 3 (shape of marginal distribution) $\times$ 2 (local dependency) $\times$ 2 (categoriality) = 12 unique data configurations. We note that a subset of these configurations, specifically the $-$D$+$C configurations, are not conceivable. This is because, since $-\text{D}$ is defined as a matrix with identical rows (each token's transition distribution is identical), both within-category and between-category similarities are 1. Hence, we have a total of 9 combinations of features, each of which generates the pretraining data. 

\Cref{ch5:tab:dists} summarizes the properties as well as the actual statistics of each of these distributions. $H(\cdot)$ measures the conditional entropy of the distribution, estimated by taking the sum of the entropies of each row $\mathbf{r} \in\mathbb{R}^{|\mathcal{V}|}$, weighted by the stationary distribution (see \Cref{ch5:appendix:entropy}). Note that the last row of each block, marked by the $-$D configuration, has a higher entropy. This is because, for the $-$D configuration, each row of the matrix is identical to each other, and each word in this distribution is i.i.d. Hence, once the shape of the marginal distribution is specified, the entropy of this matrix is automatically determined. For example, \config{Unif}{-}{-} is by definition the maximally entropic distribution over $|\mathcal{V}|$ items: $\log_2|\mathcal{V}|$.
$\text{D}_{\text{{KL}}}(\cdot\mid\mid \text{target})$ is the KL divergence between the desired marginal distribution and the actual marginal distribution of the generated matrix. We can see that the divergence is very small for all distributions. Intra-group similarities, inter-group similarities, and $\text{D}_{\text{{KL}}}(\cdot\mid\mid \text{target})$ for the \config{Dist}{-}{-} configs are 1, 1, 0, respectively, by definition.

\subsection{Results}

In \Cref{ch5:fig:exp3-ps}, we first find that no property is by itself a \textit{sufficient} condition for \gls{ih} formation when measured by \gls{ps}. For example, even under the highest bigram repetition condition (0.9-0.9), the $-$D configurations fail to produce \glspl{ih}. Similarly, local dependency appears to be necessary in this setup: no configurations with $-$D form \glspl{ih} in \Cref{ch5:fig:exp3-ps}. Intuitively, if the next token distribution is independent of the current token, the model might not be learning to utilize past tokens in context to facilitate the next token prediction. We emphasize, however, that the result varies nontrivially when using different metrics to measure the emergence of \glspl{ih}. As discussed in \Cref{appendix:results-other-metrics-exp3}, other metrics show partial \gls{ih}-like behavior in some high-repetition $-$D configurations.

Second, as we showed in \Cref{ch5:sec:exp2}, we reconfirm that some level of bigram repetition is a \textit{necessary} condition (\Cref{ch5:sec:exp2}). No configurations under the 0.1-0.1 column promote the formation of \glspl{ih}. Lastly, interestingly, marginal distribution and categoriality seem to matter only when frequency and reliability are near the decision boundary. Whereas $\pm$C and distribution shape do not affect \gls{ih} formation for 0.1-0.1 and 0.9-0.9 columns, only \config{Zipf}{+}{+} results in the formation of \glspl{ih} under the 0.1-0.3 column. The importance of skewed rank-frequency distributions, of which the Zipfian distribution is an example, is also reported in \citet{chan2022data, reddy2024the}.

\section{Emergence and Training Dynamics}\label{ch5:sec:dynamics}

\begin{figure}[t]
    \centering
    \input{figures/emergence_loss.pgf}
    \caption{Final validation loss ($y$-axis) and the emergence point as measured in \gls{ps} ($x$-axis). Colors represent the number of tokens seen per update. The solid line represents log-spaced batch sizes from 4 to 512 with the context size fixed at 1024, and the dashed line represents log-spaced context sizes from 64 to 4096 with the batch size fixed at 128. Same configurations run with three different seeds are grouped within gray shaded regions, and their centroids are used when connecting them to other points.}
    \label{ch5:fig:emergence-loss}
    \vspace{-20pt}
\end{figure}
To better understand the role of emergence points in the context of \gls{lm} pretraining dynamics in general, we plot the emergence points and final validation loss in \Cref{ch5:fig:emergence-loss}. Because models with a larger context size have an unfair advantage of larger test-time compute via attention, we feed the validation set in chunks of 64 tokens to all models. Hence, each model computes each token's loss using exactly the same amount of information regardless of the model configurations. From the solid line, which fixes the context size at 1024 and varies the batch size from 4 to 512 (dark to bright), we can see that increasing the batch size moves the emergence point earlier as we have seen in \Cref{ch5:sec:exp1}, and also makes the validation loss lower until a certain point ($B=64$) but higher beyond that point. We can also see that the emergence point does not change as much beyond $B=128$. This is reminiscent of the \gls{cbs} hypothesis \cite{mccandlish2018, merrill2026critical}: given a fixed compute, as we increase the batch size, the gain in the gradient quality diminishes, and the decrease in the number of updates results in a worse training outcome overall. In our case, we suspect that the emergence point will only become earlier with an increased batch size and better gradient update quality, and that once the gradient quality saturates, the emergence point will become constant. This V-shaped final validation loss curve is observed in the dashed line in \Cref{ch5:fig:emergence-loss} as well, which varies context size while holding batch size fixed, suggesting the importance of context size in the \gls{cbs} hypothesis.\footnote{For an analysis on the potentially \textit{later} emergence of \glspl{ih} with a \textit{larger} context size in \Cref{ch5:fig:emergence-loss}, see \Cref{appendix:retrieval}.}

A related future direction is to distinguish between a mere timing shift and a qualitatively different downstream trajectory: that is, whether models in which \glspl{ih} emerge earlier ultimately converge to similar solutions, or instead differ meaningfully in \gls{icl} behavior or other downstream measures. We leave a full characterization of such path dependence beyond the scope of the present study, which focuses on predicting when \glspl{ih} emerge. Nevertheless, the non-monotonic relationship between emergence timing and final validation loss suggests that predicting \gls{ih} emergence may be practically useful for choosing training configurations: earlier emergence is beneficial only up to a point, after which further accelerating emergence does not necessarily improve training outcomes.

\section{Discussion}\label{ch5:sec:discussion}

\subsection{Importance of task selection}

Our finding that larger context sizes promote earlier \gls{ih} emergence may appear to contrast with \citet{zucchet2025the}, who find later emergence with larger context sizes in an associative-recall task. We view these results as reflecting different effects of increasing context size. Larger contexts can simultaneously (i) increase the number of tokens per update, (ii) increase the number of repeated bigrams in natural language (\Cref{ch5:fig:nat_dist}), and (iii) make retrieval harder by increasing query-key distance and the number of distractors. Our next-token-prediction setup appears to be dominated by the first two effects, while the retrieval-difficulty effect is more directly isolated in the associative recall setup in \citet{zucchet2025the}; we discuss evidence for this effect in our setting in \Cref{appendix:retrieval}.
On the topic of task selection, it is also worth noting that subword tokenization inflates the number of repeated bigrams in natural language, and using an orthographic tokenization might lead to different results, which is a promising direction for future research.

\subsection{Implications for broader phase transition}

Phase transition is a broad term that refers to any abrupt change in a target behavior or metric of interest. It has been studied in the context of $n$-gram distributions \cite{chang-etal-2024-characterizing, chang2025bigram, michaelov2025language}, human language processing \cite{aoyama-wilcox-2025-language}, and concept spaces in vision \cite{okawa2023compositional, park2024emergence}, to name a few. Previous studies find that \glspl{ih} emerge after bigram learning \cite{olsson2022context, bietti2023birth}. Although the present study focuses on \glspl{ih}, and some aspects of the data properties may be specific to them (e.g., repetitions), a broader trend observed in \cref{ch5:sec:exp1} may generalize to other behavioral phases in \gls{lm} pretraining. For example, \citet{aoyama-wilcox-2025-language} found a similar effect of batch size on the emergence of syntactic attention structure \cite{chen2024sudden}, and understanding to what extent the size-agnostic and \gls{twu}-dependent nature of \gls{ih} emergence observed in this study also holds for other phenomena remains an important direction for future research.

\subsection{Conclusion}\label{ch5:sec:summary}

We showed that the emergence of \glspl{ih} can be predicted by batch size and context size (\Cref{ch5:sec:predict}). We also showed that the frequency and reliability of bigram repetitions can express a decision boundary for \gls{ih} formation (\Cref{ch5:sec:exp2}). Lastly, we found that, among local dependency, categoriality, the shape of the marginal distribution, frequency, and reliability, none of them alone was sufficient to ensure the formation of \glspl{ih} when measured by \gls{ps} (\Cref{ch5:sec:exp3}). We also find that local dependency coupled with high frequency and reliability always results in \gls{ih} formation, and that the shape of the marginal distribution and categoriality appear to modulate \gls{ih} formation when the frequency and reliability are near the decision boundary.
We also showed that predicting \gls{ih} emergence has practical implications for training efficiency given its non-monotonic relationship with the final validation loss under a fixed compute budget.

\subsection{Limitations}

As mentioned above, due to the limited compute resources, we only included models of up to 350M parameters in size for training, and 7B parameters for inference. Modern \glspl{lm} are orders of magnitude larger in parameter count, and it is important to test if the trend holds for larger sizes. However, given a consistent trend we observed across three orders of magnitude (ranging from 50M GPT2 to 7B Pythia), we believe that a similar trend may hold for larger models.

\section*{Impact Statement}
This paper presents work whose goal is to advance the field of Machine Learning. There are many potential societal consequences of our work, none of which we feel must be specifically highlighted here.

\section*{Acknowledgment}

We thank the 4 anonymous reviewers, whose thoughtful and insightful comments substantially improved the paper. We also thank the organizers, reviewers, and participants of the CogInterp workshop at NeurIPS 2025, where we received invaluable feedback on the earlier version of this paper.

\bibliography{trimmed_dedup}
\bibliographystyle{icml2026}

\newpage
\appendix
\onecolumn

\crefalias{section}{appendix}

\begin{table}[h]
    \centering
    \caption{List of hyperparameters used to train LMs.}
    \input{tables/hyperparameters}
    \label{tab-hyperparams}
\end{table}

We also include 6 Pythia models: 70M, 160M, 410M, 1B, 2.8B, and 6.9B. For the details of these Pythia models, see Table 1 of \citet{biderman2023pythiasuiteanalyzinglarge}.

\section{Determining the threshold for random attention}\label{appendix:mcmc}

In \Cref{ch5:fig:batch_context_rep_ps_steps} (center), the lines representing context sizes of 4, 8, and 16 seem to be somewhat flat, meaning that the model does not improve in its ability to attend back to the token necessary to complete the repeated bigram. The high \glspl{ps} associated with these models are simply due to the higher attention preceding tokens can get by chance; with the context size of 4, for example, the model has at most 4 tokens to attend back to, with the random attention of 0.25.

To systematically determine what counts as ``above random,'' we simulate the random attention over previous tokens via \gls{mcmc} using a Dirichlet distribution with a uniform prior. We find that attention weights as strong as 0.72, 0.35, and 0.16 are necessary for models with context sizes of 4, 8, and 16, respectively, to be considered \textit{above random} at the alpha level of 0.01. Hence, we conclude that these three context sizes do not promote the formation of \glspl{ih}, and in the remaining analyses in this subsection, we will focus on the rest of the models.

\section{\Cref{ch5:fig:batch_context_rep_ps_steps} plotted against the number of training tokens}\label{appendix:tokens_not_steps}
\begin{figure*}
    \centering 
    \input{figures/batch_context_rep_ps_toks.pgf}
    \caption[Developmental trajectories of \gls{ps} of \glspl{lm} with various batch sizes (left), context sizes (center), and repetitions (right) over the course of 1B tokens of pretraining.]{Developmental trajectories of \gls{ps} of \glspl{lm} with various batch sizes (left), context sizes (center), and repetitions (right) over the course of 1B tokens of pretraining, plotted against the number of updates (a) and the number of tokens (b). BS, CS, \%R stand for batch size, context size, and the proportion of chunks with natural bigram repetitions, respectively.}
    \label{ch5:fig:batch_context_rep_ps_toks}
\end{figure*}

In \Cref{ch5:fig:batch_context_rep_ps_steps}, we plotted how batch size, context size, and bigram repetition rate affect the formation point of \glspl{ih}, measured in the number of training steps. Trivially, models with different batch sizes and context sizes will have been exposed to different numbers of tokens at the same training step. This could raise the question of whether or not the observed shifting effect is just an artifact of the total number of training tokens being different on the same point on the $x$-axis. Hence, we show the same graph plotted against the number of total training tokens, instead of the number of training steps in \Cref{ch5:fig:batch_context_rep_ps_toks}. Here again, we observe the shifting effect, but in the opposite direction: the smaller the batch/context size, the earlier the phase transition point. Therefore, we can say that, even when measured in the number of training tokens, \glspl{ih} form at different points when we change batch size and/or context size, ruling out the possibility that the observed shifting effect in \cref{ch5:fig:batch_context_rep_ps_steps} is an artifact of each point on the $x$-axis representing a different number of pretraining \textit{tokens}.

\section{Fitting procedure for the prediction of emergence points}\label{appendix:fitting-procedure}
Here, we provide details on the procedure for fitting a regression model to predict emergence points.

\subsection{Emergence points}\label{appendix:fitting-procedure-emergence}
First, given a metric of interest that goes through an abrupt change, we need a way to systematically identify the point (in number of training steps or tokens) at which it is said to go through a phase transition point (or a certain capability measured by the score ``emerges''). Prior work has used a few different approaches, such as the sharpest slope \cite{chen2024sudden} and the first point at which the metric exceeds a given threshold \cite{aoyama-wilcox-2025-language}. Here, given a curve that shows the development of a given metric's score over the course of some time metric, such as training steps and tokens (e.g., \cref{ch5:fig:batch_context_rep_ps_toks}), we fit a \gls{pwlf} model to the curve with three segments. We choose three as we observe the initial stagnation phase, where the score is around 0, the abrupt improvement phase, and the eventual plateau phase. We then take the ``knot'' between the first two segments and operationalize that point as the emergence point.

\subsection{Data for fitting the linear regression models}\label{appendix:fitting-procedure-data}
The aforementioned emergence identification assigns a single emergence point for each model. Hence, we have as many data points as there are models. We have trained GPT2-50M models with 18 different batch and context size configurations (\Cref{ch5:sec:exp1}), as well as GPT2-125M and GPT2-350M with 7 different configurations each. However, because we have shown that emergence was not observed in the context sizes of \{4, 8, 16\}, we exclude those models in this analysis. We finally add 6 Pythia models \{70M, 160M, 410M, 1B, 2.8B, 6.9B\} all trained on batch size of 1024 and context size of 2048. In total, we fit a regression model on these 35 models. We reported the ``fit'' of this regression model trained on all of the data points; however, we also report the results from cross-fold validation as well in \Cref{tab:cv}.

\subsection{Regression model details for phase transition prediction}\label{appendix:fitting-procedure-regression}

In \cref{ch5:sec:exp1}, we fitted a linear regression in log space using batch size, context size, and model size as predictors for the phase transition point. 

\Cref{tab:full_regression} summarizes the linear regression model with all three predictors. A few consistent trends emerge: first, batch size and context size are (1) statistically significant across the board, regardless of the metric used to define the phase transition point, and (2) in the negative direction, meaning that the increase in these predictors leads to \textit{earlier} emergence. Second, model size is non-significant across the board, regardless of the metric. Hence, we conclude that the emergence points are independent of model size. The final regression models without the model size are summarized in \Cref{tab:nomodelsize_regression}.

Lastly, as mentioned earlier, we also report the results of 5-fold cross-validation of the full regression models and the models without model size in \Cref{tab:cv}. Each fold trains the regression model on 80\% of the training data (28 models) and predicts the emergence points of the remaining 20\% of the data (7 models). Most test $R^2$s range from 0.8 to 0.9, indicating an excellent fit to the \textit{held-out} data points. Note that $R^2_{test}$ is computed as:
\begin{equation}
    1-\frac{SS_{residual}}{SS_{total}}
\end{equation}
and the value of 1 indicates a perfect prediction on the test set.

\begin{table}[t]
    \centering
    \caption{A summary of linear regression analyses with an intercept, logB (batch size), logC (context size), and logM (model size) as independent variables and the phase transition point (in the number of updates) in each of the four metrics, PS, LA, AR (accuracy), AR (mean rank), as dependent variables. Significantly \textbf{\textcolor{mdgreen}{positive}} and \textbf{\textcolor{mdred}{negative}} test statistics are colored in \textbf{\textcolor{mdgreen}{green}} and \textbf{\textcolor{mdred}{red}}, respectively.}
    \input{tables/full_regression}

    \label{tab:full_regression}
\end{table}

\begin{table}[t]
    \centering
    \caption{A summary of linear regression analyses with an intercept, logB (batch size), and logC (context size) as independent variables and the phase transition point (in the number of updates) in each of the four metrics, PS, LA, AR (accuracy), AR (mean rank), as dependent variables. Significantly \textbf{\textcolor{mdgreen}{positive}} and \textbf{\textcolor{mdred}{negative}} test statistics are colored in \textbf{\textcolor{mdgreen}{green}} and \textbf{\textcolor{mdred}{red}}, respectively.}
    \input{tables/nomodelsize_regression}

    \label{tab:nomodelsize_regression}
\end{table}

\begin{table}[t]
    \centering
    \caption{Cross-fold validation results for the full regression model (left) and the regression model without model size (right). Each fold trains the regression model on 80\% of the data and predicts on the remaining 20\% of the data. Recall that $\beta$, $\gamma$, $\theta$ are coefficients of batch size, context size, and model size, respectively. Prediction quality is measured in $R^2_{test}=\frac{SS_{residual}}{SS_{total}}$}
    \input{tables/cv}
    \label{tab:cv}
\end{table}

\clearpage

\section{Variability across random seeds}\label{appendix:var}

We train a subset of 6 batch-context configurations with 2 additional random seeds to quantify the variability across 3 random seeds. \Cref{tab:var} (left) summarizes the mean \gls{ih} emergence points, as measured in the number of updates $U_{PT}$, and their standard deviations. \Cref{tab:var} (right) summarizes the mean fitted parameter values for the predictive law described in \Cref{ch5:eq:general}: $U_{\textsc{pt}}= e^\alpha B^\beta C^\gamma$.

\begin{table}[h]
    \centering
    \caption{Uncertainty estimation of emergence points (left) and predictive law parameters (right). Configurations are presented in (batch, context) notation.}
    \input{tables/var}
    \label{tab:var}
\end{table}

\section{Distribution of bigram repetitions with various context sizes}\label{appendix:nat-rep}

\begin{figure}[h]
\centering
\input{figures/abab_dist.pgf}
    \caption[Smoothed distribution of chunks with various numbers of bigram repetitions.]{Smoothed distribution of chunks with various numbers of bigram repetitions. Context sizes 1024 and 2048 are not visible, and hence removed from the plot. The plot is truncated at y=0.6 for readability, but context sizes of 4, 8, and 16 had >95\% of chunks with no bigram repetitions.}
    \label{ch5:fig:nat_dist}
\end{figure}

Since adding or removing bigrams in naturally occurring texts introduces noise, such as broken syntax, we manipulate the frequency of repeated bigrams in natural language data by first putting tokenized texts into chunks of $c$, where $c$ is the context size of the \gls{lm}, and then selecting those natural chunks to ensure $p\%$ of the chunks of the resulting training data include no bigram repetition at all, where $p$ is the parameter we can control. Modern \glspl{lm} have a context size of at least 1024. However, in naturally occurring texts, sequences of 1024 tokens without any repeated bigrams \abab{A}{B} in them are very rare, if not non-existent. In general, as shown in \Cref{ch5:fig:nat_dist}, larger context size trivially tends to contain more repeated bigrams, and we find that the context size of 64 strikes a balance between including enough context and containing a good number of chunks with and without repeated bigrams. Hence, in \Cref{ch5:sec:exp2} and \Cref{ch5:sec:exp3}, we use \glspl{lm} with a context size of 64.

\section{Frequency \paba{A}{B} and Reliability \pbgivenaba{A}{B}}\label{appendix:freq_and_rel}

Recall that \citet{elhage2021mathematical, olsson2022context} define \glspl{ih} as specific heads that complete the repeated \abab{A}{B} sequence when seeing \aba{A}{B}. As a naive hypothesis, we speculate that the presence of repeated bigrams (separated by an arbitrarily long sequence of tokens within the LM's context size) is essential for \gls{ih} formation. Consider the following sequence, which we will use as a running example:
\begin{table}[ht]
    \centering
    \caption{$R_U$ and $R_B$ of an example sequence. This sequence has a $P(R_U)=\frac{4}{10}$ and $P(R_B)=\frac{2}{10}$, hence \textbf{frequency} and \textbf{reliability} are 0.4 and 0.5, respectively.}
    \input{tables/abab_example}
    \label{ch4:ex:abab}
\end{table}
\noindent In this example, there are 4 tokens that occur more than once (i.e., \tone, \ttwo, \tfive, \tsix). For the first occurrence of these 4 tokens, each of them is followed by \textit{something}; for example, \tone is followed by \ttwo, and \ttwo is followed by \tthree. For the second occurrence of these 4 tokens, one can induce that a bigram tends to repeat \textit{in-context}, if these 4 tokens are reliably followed by what followed them the first time they occurred; namely, if \tone, \ttwo, \tfive, \tsix, are followed by \ttwo, \tthree, \tsix, \tone, respectively. In \Cref{ch4:ex:abab}, there exist 4 such opportunities for learning the bigram repetition, and 2 such opportunities actually reward such learning, since only the second occurrences of \tone and \tfive are followed by the same bigram continuations of their first occurrences. Here, we have touched upon the two knobs we aim to define here. \textbf{Frequency} involves the tokens that are of the word type occurring for the $n$-th time where $n\ge 2$ (\tone, \ttwo, \tfive, \tsix in \Cref{ch4:ex:abab}). \textbf{Reliability} measures, of all such tokens, how many of them actually complete the same bigram continuation (\ttwo, \tfive in \Cref{ch4:ex:abab}). We will now define each of these two measures formally.

\noindent\textbf{Frequency}: given a binary variable \textit{unigram repetition} $R_U\in\{0,1\}$, whose value is 1 if a given token is the second occurrence of \tok{A} of the \aba{A}{B} sequence (i.e., a repeated unigram), and 0 otherwise, we define ``frequency'' as $P(R_U)$. Informally, in induction terms, this is $\paba{A}{B}$, or the rate at which \aba{A}{B} sequence is encountered. In practice, this is equivalent to the sum of all unigram counts in a given sequence, with the first occurrence removed:
\begin{equation}\label{ch5:eq:frequency}
    \paba{A}{B}=\frac{1}{|\mathbf{s}|}\sum_{w\in \mathcal{V}}\max(c(w, \mathbf{s})-1, 0)
\end{equation}
where $\mathcal{V}$ is the vocabulary, or the set of all possible unigrams, and $c(w, \mathbf{s})$ is a count of a word $w$ in a sequence $\mathbf{s}$ of the same size as the model's context size. In \Cref{ch4:ex:abab}, we saw 4 such tokens (tokens where $R_U=1$: \textbf{\tone}, \textbf{\ttwo}, \textbf{\tfive}, \textbf{\tsix}) in the context of size 10, hence $P(R_U)=\paba{A}{B}=\frac{4}{10}$. Technically, consecutive occurrences of a given token type should not be counted; however, this does not happen frequently in the real data, and no tokens were allowed to occur twice in a row in synthetic data, trivializing this problem (see \Cref{ch5:alg:corpus}).

\noindent\textbf{Reliability}: given a binary variable \textit{bigram repetition} $R_B\in\{0,1\}$, whose value is 1 if a given token's \textit{next token} is the second occurrence of \tok{B} of the \abab{A}{B} sequence (i.e., a repeated bigram), and 0 otherwise, we define ``reliability'' as $\frac{P(R_B)}{P(R_U)}$. Informally, in induction term, $P(R_B)$ is $\pabab{A}{B}$, and reliability, or $\frac{P(R_B)}{P(R_U)}$, is equivalent via the chain rule to the conditional probability \pbgivenaba{A}{B}: the rate at which an \aba{A}{B} sequence is followed by \tok{B}:

\begin{equation}
    \label{ch5:eq:reliability}
    \pbgivenaba{A}{B}=\frac{\pabab{A}{B}}{\paba{A}{B}}=\frac{P(R_B)}{P(R_U)}=\frac{\frac{1}{|\mathbf{s}|}\sum_{b\in \mathcal{B}}\max(c(b, \mathbf{s})-1, 0)}{\frac{1}{|\mathbf{s}|}\sum_{w\in \mathcal{V}}\max(c(w, \mathbf{s})-1, 0)}
\end{equation}
where $\mathcal{B}$ denotes the set of all possible bigrams, and $c(b, \mathbf{s})$ the count of the bigram $b$ in a given sequence $\mathbf{s}$. In \Cref{ch4:ex:abab}, of all the 10 positions (tokens), 2 completed the \abab{A}{B} sequence (tokens where $R_B=1$: second occurrences of \ttwo and \tsix), hence $P(R_B)=\pabab{A}{B}=\frac{2}{10}$. Therefore, reliability is $\frac{2}{10}\div\frac{4}{10}=\frac{1}{2}$. It might make more intuitive sense to compute this directly without using the chain rule: of all the 4 tokens that complete the \aba{A}{B} sequence, 2 of them are followed by \tok{B}, hence $\frac{2}{4}=\frac{1}{2}$. Equivalently, of the tokens where $R_U=1$ in \Cref{ch4:ex:abab}, half of them also have $R_B=1$. However, for computational purposes, the chain rule is much simpler, which is the reason we introduced the chain rule-based calculation above. In this study, ``frequency'' and \paba{A}{B} are used interchangeably, and so are ``reliability'' and \pbgivenaba{A}{B}.

Note that this is a more precise characterization of what \citet{chan2022data} called ``burstiness.'' In their formulation, where the task was to predict the label of an image given image-label pairs in-context, a ``bursty'' sequence contains certain image-label pairs more often than others, while controlling for the marginal distribution over all sequences. This is effectively equivalent to increasing both frequency and reliability in our terms. Two sequences of image-label pairs ($\mathsf{I_i,L_i}$): $\langle\mathsf{I_1,L_1,I_2,L_2,I_3,L_3,I_4,L_4}\rangle$ (non-bursty) and $\langle\mathsf{I_1,L_1,I_2,L_2,I_1,L_1,I_3,L_3}\rangle$ (bursty) can be reformulated as $\langle \mathsf{A,B,C,D,E,F,G,H}\rangle$ and $\langle \mathsf{A,B,C,D,A,B,E,F}\rangle$, respectively, and the former has frequency and reliability of 0, whereas the latter has frequency and reliability of 1/8 and 1/2 (\aba{A}{B} is followed by \tok{B}, but \aba{B}{C} is not followed by \tok{C}), respectively.
\begin{figure}[t]
\centering
    \input{figures/abab_probas.pgf}
    \caption[Distribution of chunks with various \paba{A}{B} and \pbgivenaba{A}{B} for each context size.]{Distribution of chunks with various \paba{A}{B} and \pbgivenaba{A}{B} for each context size. Only the quartile box, median (center line in each box), mean (diamond), and whiskers are shown, and outliers are not shown for readability.}
    \label{ch5:fig:abab_probas}
\end{figure}
We can verify that natural texts with various chunk sizes are not only different in terms of the number of repeated bigrams (as shown in \Cref{ch5:fig:nat_dist}), but also in the two probabilities defined above. \Cref{ch5:fig:abab_probas} confirms this: both \paba{A}{B} and \pbgivenaba{A}{B} increase as the context size increases. While \paba{A}{B} seems to increase log-linearly ($x$-axis is on a log scale), the increase in \pbgivenaba{A}{B} seems to slow down. Since \glspl{ih} were not forming, or very weak at most, for a context size of 32, we speculate that the threshold values of \paba{A}{B} and \pbgivenaba{A}{B} are somewhere between 0.1--0.2, and 0.1--0.15, respectively.


\section{Algorithm for Frequency- and Reliability-Constrained Training Data Generation}\label{appendix:algo}

\input{tables/algo}

To fully control the two properties, frequency \paba{A}{B} and reliability \pbgivenaba{A}{B} in text data, we need a way to sample words from some distribution, while enforcing desired values of these two knobs. To this end, we approximate natural language by first tokenizing texts from the English subcorpus of the Common Crawl Corpus (\texttt{CC100}; \citealp{conneau-etal-2020-unsupervised, wenzek-etal-2020-ccnet}) and collecting token bigram statistics. We then create a token-to-token transition matrix $T\in \mathbb{R}^{|\mathcal{V}|\times|\mathcal{V}|}$, where $\mathcal{V}$ is the vocabulary. We use an off-the-shelf GPT2 tokenizer, but reduce the vocabulary size to 10,000.

In \Cref{ch5:alg:corpus}, we outline the semi-synthetic data generation algorithm, where $(\mathcal{A}\space\backslash\space \mathcal{B})$ denotes an asymmetric difference, or a set of elements in $\mathcal{A}$ but not in $\mathcal{B}$, $\{x \mid x\in \mathcal{A}\text{ and } x \notin \mathcal{B}\}$. $\mathcal{U}_{<t-1}$ is a set of unigrams attested before time step $t-1$, and $\mathcal{B}_{x}$ denotes a set of attested bigram continuations of token $x$.
The idea is that we first generate the first half of the sequence by randomly walking through the Markov chain. This is to ensure that a sufficient number of unique tokens are present in the sequence before the restricted generation can take place; otherwise, \Cref{ch5:alg:corpus} often produces degenerate sequences especially with high values of \paba{A}{B} and \pbgivenaba{A}{B}. For the second half, based on the condition (if the prefix constitutes \aba{A}{B}) and constraints (to make a new \aba{A}{B} and/or \abab{A}{B}), the word at each time step $w_t$ is sampled from the distribution $\mathcal{D}$ restricted to some subset $\mathcal{S}$ that satisfies the conditions and constraints. For simplicity, no tokens were allowed to occur consecutively.

Note that the constraints \paba{A}{B} and \pbgivenaba{A}{B} are imposed at each token generation step uniformly, meaning that we do not devise a token-specific repetition rate. It is an open question whether or not this is a sensible simulation of the token distribution in natural language: at its core, it comes down to the question of to what extent each word in natural language differs in its probability of participating in repeated bigrams beyond their differences in unigram and bigram distributions, and we will leave this to future work.

\section{Matrix Optimization}\label{ch5:appendix:optimization}

In \Cref{ch5:sec:exp3}, we defined the underlying Markov Processes as transition matrices, which were then used to generate synthetic data, via a set of desired properties. These transition matrices were optimized using the Adam optimizer to satisfy the desired properties as closely as possible.
\begin{equation}    \mathcal{L}=\lambda_1\mathcal{L}_{D}+\lambda_2\mathcal{L}_{E}+\lambda_3\mathcal{L}_{P}+\lambda_4\mathcal{L}_{WC}+\lambda_5\mathcal{L}_{WA}
\end{equation}
where the 5 terms correspond to distribution loss, entropy loss, peakedness loss, within-category loss, and across-category loss, respectively. The 5 $\lambda$s correspond to the weighting factors.

\noindent\textbf{Distribution loss ($\mathcal{L}_{D}$).} This term is to ensure the transition matrix $T\in\mathbb{R}^{|\mathcal{V}|\times|\mathcal{V}|}$ has a marginal distribution of the desired shape (Uniform, Gaussian, or Zipfian, as discussed in \Cref{ch5:sec:exp3}). We penalize the divergence from the desired shape by including KL divergence between the actual marginal distribution and the desired distribution:
\begin{equation}
\mathcal{L}_{D}=\sum_{i=1}^{|\mathcal{V}|}P(w_i)\log\frac{P(w_i)}{Q(w_i)},
\end{equation}
where P and Q are desired and actual marginal distributions, respectively.

\noindent\textbf{Entropy loss ($\mathcal{L}_{E}$).} This term is to ensure each transition matrix is comparable in predictability. Because a transition matrix sampled from natural language had the entropy of $\approx$6.2, we set the target entropy value to be 6.2, and included the squared difference as a loss term:
\begin{equation}
\mathcal{L}_{E}=\|H_{\text{target}}-H(T)\|_2^2,
\end{equation}
where the estimation of $H(T)$ is detailed in \Cref{ch5:appendix:entropy}.

\noindent\textbf{Peakedness loss ($\mathcal{L}_{P}$).} We find that the matrix optimization often suffers from a degenerate matrix, where the desired properties are satisfied by allocating a very large probability mass to a single token in each row. To mitigate this problem, we include the mean of row-wise maximum probabilities:
\begin{equation}
\mathcal{L}_{P}=\frac{1}{|\mathcal{V}|}\sum_{i=1}^{|\mathcal{V}|}\max\space T_{i,:},
\end{equation}
where $T_{i,:}$ is the $i$-th row of the matrix $T$.

\noindent\textbf{Within-category loss ($\mathcal{L}_{WC}$) and across-category loss ($\mathcal{L}_{AC}$).} As defined in \Cref{ch5:sec:exp3}, the within-category similarity is the mean similarity of all pairs of words within a category, whereas the across-category similarity is the mean similarity of all pairs of words from different categories. For the $+$C configuration, we set the target within- and across-category similarities to be 0.4 and 0.1, respectively, and both were set to be 0.1 for the $-$C configuration. The squared differences between the actual and desired within-/across-category similarities were included as loss terms. The within-category loss is defined as:
\begin{equation}
\mathcal{L}_{WC}=\|\text{WC}_{\text{target}}-\text{WC}(T)\|_2^2,
\end{equation}
where WC is within-category similarity:
\begin{equation}
    \frac{1}{N}\sum_{c\in C}\sum_{w, w'\in c,w\ne w'}{sim(w,w')}
\end{equation}
Similarly, across-category loss is defined as:
\begin{equation}
\mathcal{L}_{AC}=\|\text{AC}_{\text{target}}-\text{AC}(T)\|_2^2,
\end{equation}
where AC is across-category similarity:
\begin{equation}
    \frac{1}{N}\sum_{c, c'\in C, \space c\ne c'}\sum_{w\in c}\sum_{w'\in c'}{sim(w,w')}
\end{equation}

We optimize the matrix with these loss terms for 5,000 steps with $\lambda_1=100$, $\lambda_2=0.01$, $\lambda_3=0.1$, $\lambda_4=\lambda_5=5$. 

\section{Conditional Entropy}\label{ch5:appendix:entropy}
Conditional entropy is defined as:
\begin{equation}
    H(X \mid Y) = \sum_{y \in Y} P(y) \left[ \sum_{x \in X} P(x \mid y) \log \frac{1}{P(x \mid y)} \right]
\end{equation}
For a transition matrix $T\in\mathbb{R}^{|\mathcal{V}|\times|\mathcal{V}|}$, where $\mathcal{V}$ is the vocabulary, it can be expressed as:
\begin{equation}\label{ch5:appendix:eq:condent}
    H(T) = \sum_{i=1}^{|\mathcal{V}|} P(w_i) \left[ \sum_{j=1}^{|\mathcal{V}|} P(w_j \mid w_i) \log \frac{1}{P(w_j \mid w_i)} \right]
\end{equation}
Since the transition matrix $T$ is a row-stochastic matrix, and each row sums to 1, the conditional probability $P(w_j|w_i)$ is an entry $T[i,j]$. The marginal probability $P(w_i)$ can be estimated by obtaining the stationary distribution of the transition distribution $T$, which is a left eigenvector with the eigenvalue of 1. Assume a ground-truth stationary distribution $\pi$; this stationary distribution, which is the unigram distribution the transition matrix converges to, should remain unchanged after transitions:
\begin{align}
    \pi T&=\pi
\end{align}
An eigenvector is a vector that only gets scaled by a factor $\lambda$ after a linear transformation $L$. Hence, we can find the stationary unigram distribution $\pi$ by finding the left eigenvector with the eigenvalue of 1 of a linear transformation $T$.

Now that we have the stationary distribution $\pi$, \Cref{ch5:appendix:eq:condent} can be expressed as a matrix multiplication: 
\begin{equation}
H(T)=\sum_{i=1}^{|\mathcal{V}|}\pi[i]\cdot H(T[i,])
\end{equation}

\section{Context size and retrieval difficulty}\label{appendix:retrieval}

\begin{figure}[t]
    \centering
    \input{figures/emergence_loss.pgf}
    \caption{\Cref{ch5:fig:emergence-loss} reproduced for readability. Final validation loss ($y$-axis) and the emergence point as measured in \gls{ps} ($x$-axis). Colors represent the number of tokens seen per update. The solid line represents log-spaced batch sizes from 4 to 512 with the context size fixed at 1024, and the dashed line represents log-spaced context sizes from 64 to 4096 with the batch size fixed at 128. Same configurations run with three different seeds are grouped within gray shaded regions, and their centroids are used when connecting them to other points.}
    \label{appendix:fig:emergence-loss}
\end{figure}

In \Cref{ch5:sec:dynamics}, we mainly focus on the observation that both batch size and context size participate in the V-shaped validation loss curve (\Cref{appendix:fig:emergence-loss}). However, the picture with the context size seems more complicated. Unlike batch size, where an increase always results in an earlier emergence point, increasing context size could potentially make the emergence later. This is due to an interplay of (at least) 3 effects of increasing the context size: (1) more tokens seen per update (``shifting'' in \Cref{ch5:sec:exp1}), (2) more repetitions per sequence (``slanting'' in \Cref{ch5:sec:exp1}), and (3) higher retrieval difficulty. The \gls{cbs} hypothesis suggests that the improvement in gradient quality in (1) saturates, and we suspect that the increase in frequency and reliability in  (2) also does (e.g. increase in \pbgivenaba{A}{B} slows down with larger context sizes in \Cref{ch5:fig:nat_dist}). \citet{zucchet2025the} show that the third effect, the retrieval difficulty, increases with a larger context size, which delays the emergence. We surmise that, with the saturation of (1) and (2), the effect (3) dominates, which could result in a later emergence point in \gls{lm} pretraining. However, we emphasize that this explanation is only a plausible account of the observed trend: it remains unclear whether the later emergence at large context sizes observed in \Cref{appendix:fig:emergence-loss} is indeed caused by retrieval difficulty, or whether this reversal is systematic across settings. If this account is correct, the predictive law presented in \Cref{ch5:sec:predict} may primarily describe the pre-saturation regime of effects (1) and (2), before increased retrieval difficulty in (3) begins to dominate. A precise characterization of this interplay remains an important future work.

\section{Analyses with other metrics}\label{appendix:results-other-metrics}
\subsection{Experiment 1}\label{appendix:results-other-metrics-exp1}
We replicate the analyses from \Cref{ch5:sec:exp1} in other metrics: \gls{la} in \Cref{ch5:fig:batch_context_rep_la_steps} and \Cref{ch5:fig:batch_context_rep_la_toks}, \gls{ar} (accuracy) in \Cref{ch5:fig:batch_context_rep_acc_steps} and \Cref{ch5:fig:batch_context_rep_acc_toks}, and \gls{ar} (mean rank) in \Cref{ch5:fig:batch_context_rep_mean_rank_steps} and \Cref{ch5:fig:batch_context_rep_mean_rank_toks}. Notably, the shifting and slanting effects discussed in \Cref{ch5:sec:exp1} are visibly captured by \gls{la} (\Cref{ch5:fig:batch_context_rep_la_steps}) and by \gls{ar} (mean rank; \Cref{ch5:fig:batch_context_rep_mean_rank_steps}), but not as pronounced in \gls{ar} (accuracy; \Cref{ch5:fig:batch_context_rep_acc_steps}). This is presumably, again, due to the discrete nature of accuracy, where the metric only captures whether the model ranks the target token as the most probable token or not. Another point worth noting is that, \gls{ar} (accuracy) gives a false impression that a batch size of 512 fails in the associative recall task, as shown by the yellow curve in \Cref{ch5:fig:batch_context_rep_la_steps}. However, when looking at the mean rank as opposed to accuracy, we can see that it goes through an abrupt drop (improvement) in the model trained with a batch size of 512 (\Cref{ch5:fig:batch_context_rep_mean_rank_steps}), highlighting the importance of analyzing multiple metrics.
\begin{figure*}[h] 
  \begin{subfigure}[b]{\linewidth}
    \centering
    \makebox[\linewidth][l]{\textbf{(a)}} \\
    \input{figures/batch_context_rep_la_steps.pgf}
    \phantomcaption 
    \label{ch5:fig:batch_context_rep_la_steps}
\end{subfigure}
  \begin{subfigure}[b]{\linewidth}
    \centering 
    \makebox[\linewidth][l]{\textbf{(b)}} \\
    \input{figures/batch_context_rep_la_toks.pgf}
    \phantomcaption 
    \label{ch5:fig:batch_context_rep_la_toks}
  \end{subfigure}
  \vspace{-0.5cm}
    \caption[Developmental trajectories of \gls{la} of \glspl{lm} with various batch sizes (left), context sizes (center), and repetitions (right) over the course of 1B tokens of pretraining.]{Developmental trajectories of \gls{la} of \glspl{lm} with various batch sizes (left), context sizes (center), and repetitions (right) over the course of 1B tokens of pretraining, plotted against the number of updates (a) and the number of tokens (b). BS, CS, \%R stand for batch size, context size, and the proportion of chunks with natural bigram repetitions, respectively.}
\end{figure*}

\begin{figure*}[h] 
  \begin{subfigure}[b]{\linewidth}
    \centering
    \makebox[\linewidth][l]{\textbf{(a)}} \\
    \input{figures/batch_context_rep_acc_steps.pgf}
    \phantomcaption 
    \label{ch5:fig:batch_context_rep_acc_steps}
\end{subfigure}
  \begin{subfigure}[b]{\linewidth}
    \centering 
    \makebox[\linewidth][l]{\textbf{(b)}} \\
    \input{figures/batch_context_rep_acc_toks.pgf}
    \phantomcaption 
    \label{ch5:fig:batch_context_rep_acc_toks}
  \end{subfigure}
  \vspace{-0.5cm}
    \caption[Developmental trajectories of \gls{ar} (accuracy) of \glspl{lm} with various batch sizes (left), context sizes (center), and repetitions (right) over the course of 1B tokens of pretraining.]{Developmental trajectories of \gls{ar} (accuracy) of \glspl{lm} with various batch sizes (left), context sizes (center), and repetitions (right) over the course of 1B tokens of pretraining, plotted against the number of updates (a) and the number of tokens (b). BS, CS, \%R stand for batch size, context size, and the proportion of chunks with natural bigram repetitions, respectively.}
\end{figure*}

\begin{figure*}[ht] 
  \begin{subfigure}[b]{\linewidth}
    \centering
    \makebox[\linewidth][l]{\textbf{(a)}} \\
    \input{figures/batch_context_rep_mean_rank_steps.pgf}
    \phantomcaption 
    \label{ch5:fig:batch_context_rep_mean_rank_steps}
\end{subfigure}
  \begin{subfigure}[b]{\linewidth}
    \centering 
    \makebox[\linewidth][l]{\textbf{(b)}} \\
    \input{figures/batch_context_rep_mean_rank_toks.pgf}
    \phantomcaption 
    \label{ch5:fig:batch_context_rep_mean_rank_toks}
  \end{subfigure}
  \vspace{-0.5cm}
    \caption[Developmental trajectories of \gls{ar} (mean rank) of \glspl{lm} with various batch sizes (left), context sizes (center), and repetitions (right) over the course of 1B tokens of pretraining.]{Developmental trajectories of \gls{ar} (mean rank) of \glspl{lm} with various batch sizes (left), context sizes (center), and repetitions (right) over the course of 1B tokens of pretraining, plotted against the number of updates (a) and the number of tokens (b). BS, CS, \%R stand for batch size, context size, and the proportion of chunks with natural bigram repetitions, respectively.}
\end{figure*}

\clearpage

\subsection{Experiment 2}\label{appendix:results-other-metrics-exp2}

We replicate the analyses from \Cref{ch5:sec:exp2} with \gls{la} in \Cref{ch5:fig:nat_grid_la}, \gls{ar} (accuracy) in \Cref{ch5:fig:nat_grid_acc}, and \gls{ar} (mean rank) in \Cref{ch5:fig:nat_grid_mr}. Overall, the decision boundaries found in these results are consistent across different metrics. For example, \gls{ps} and \gls{la} seem to draw the same line between the configurations where \glspl{ih} emerge and those where \glspl{ih} do not emerge. However, it is important to note that, while the bottom right region, $\in\{\paba{A}{B}>0.1 \cap \pabab{A}{B} > 0.5\}$, is filled with highly strong \glspl{ih} (as indicated by bright yellow) for \gls{ps}, it is not the case for \gls{la}. In particular, the bottom right configuration, \paba{A}{B}=\pabab{A}{B}=0.9, results in a moderate \gls{ih}, weaker than other configurations with less repetitions.

\gls{ar} (accuracy), on the other hand, seems not to show a line as clear as the other three metrics. While the no-\gls{ih} region seems to be very similar, \gls{ih}-regions near the decision boundary in \gls{ps} and \gls{la} are more graded for \gls{ar} (accuracy). This indicates that even with the emergence of strong \glspl{ih} high in both \gls{ps} and \gls{la}, \gls{ar} accuracy can still be low. This again highlights the sequential nature of the emergence of latent and surface capabilities \cite{reddy2024the}. 

\begin{figure*}[h] 
  \begin{subfigure}[b]{0.5\linewidth}
    \centering
    \makebox[\linewidth][l]{\textbf{(a)}} \\
    \input{figures/grid_ps.pgf}
    \phantomcaption 
    \label{ch5:fig:nat_grid_ps}
\end{subfigure}
  \begin{subfigure}[b]{0.5\linewidth}
    \centering 
    \makebox[\linewidth][l]{\textbf{(b)}} \\
    \input{figures/grid_la.pgf}
    \phantomcaption 
    \label{ch5:fig:nat_grid_la}
  \end{subfigure}
    \caption[Best \gls{ps} (a) and best \gls{la} (b) across all heads at the end of the training for each frequency reliability combination.]{Best \gls{ps} (a) and best \gls{la} (b) across all heads at the end of the training for each frequency reliability combination. Scores are represented in colors, with brighter colors representing higher scores.}
\end{figure*}
\begin{figure*}[h] 
  \begin{subfigure}[b]{0.5\linewidth}
    \centering
    \makebox[\linewidth][l]{\textbf{(a)}} \\
    \input{figures/grid_acc.pgf}
    \phantomcaption 
    \label{ch5:fig:nat_grid_acc}
\end{subfigure}
  \begin{subfigure}[b]{0.5\linewidth}
    \centering 
    \makebox[\linewidth][l]{\textbf{(b)}} \\
    \input{figures/grid_mean_rank.pgf}
    \phantomcaption 
    \label{ch5:fig:nat_grid_mr}
  \end{subfigure}
    \caption[\gls{ar} accuracy (a) and \gls{ar} mean rank (b) at the end of the training for each frequency reliability combination.]{\gls{ar} accuracy (a) and \gls{ar} mean rank (b) at the end of the training for each frequency reliability combination. Scores are represented in colors, with brighter colors representing higher scores.}
\end{figure*}

\clearpage
\subsection{Experiment 3}\label{appendix:results-other-metrics-exp3}

We showed in \Cref{ch5:sec:exp3} that, when measured by \gls{ps}, \glspl{ih} form when the underlying distribution of the data generation process is $+$D and high in both repetition frequency and reliability (0.9-0.9) regardless of the distribution shape, and when it is $+$D$+$C and Zipfian when the repetition frequency and reliability are near the decision boundary (0.1-0.3). Here, we replicate the same analyses in \gls{la} (\Cref{ch5:fig:exp3-la}), \gls{ar} (accuracy; \Cref{ch5:fig:exp3-acc}), and \gls{ar} (mean rank; \Cref{ch5:fig:exp3-mr}). We make two observations. First, in each metric, \config{Zipf}{+}{+} is consistently the only distribution shape that promotes the emergence of \glspl{ih} when the bigram repetition is near the decision boundary (0.1-0.3).

Second, and more importantly, contrary to the observation we made in \Cref{ch5:sec:exp3}, \glspl{ih} do form even when the underlying distribution is $-D$, meaning that each token is i.i.d. However, it is important to note that the training data sample generated from this distribution is highly unlikely to be $-$D, and the tokens in the training data are not i.i.d., given the constrained generation process described in \Cref{appendix:algo}. Because we impose \paba{A}{B} and \pabab{A}{B} while sampling from the generation, samples are not independent of one another, violating the $-$D constraint. However, the reason why these training data sampled from underlying distributions \config{*}{-}{-} with high bigram repetition (0.9-0.9) lead to a low score in \gls{ps} (\Cref{ch5:fig:exp3-ps}) but high scores in other metrics (\Cref{ch5:fig:exp3-la,ch5:fig:exp3-acc,ch5:fig:exp3-mr}), remains unclear, and warrants further investigation. Generally, studying these heads that are low in \gls{ps} but high in \gls{la} could potentially reveal an alternative mechanism for \glspl{ih}.
\begin{figure*}[h] 
  \begin{subfigure}[b]{0.5\textwidth}
    \centering
    \makebox[\linewidth][l]{\textbf{(a)}} \\
    \input{figures/exp3_ps.pgf}
    \phantomcaption 
    \label{ch5:fig:exp3-ps-2}
\end{subfigure}
  \begin{subfigure}[b]{0.5\textwidth}
    \centering 
    \makebox[\linewidth][l]{\textbf{(b)}} \\
    \input{figures/exp3_la.pgf}
    \phantomcaption 
    \label{ch5:fig:exp3-la}
  \end{subfigure}
    \caption[Best \gls{ps} (a) and best \gls{la} (b) at the end of the training for each frequency reliability combination.]{Best \gls{ps} (a) and best \gls{la} (b) at the end of the training for each frequency reliability combination. Scores are represented in colors, with brighter colors representing higher scores.}
\end{figure*}
\begin{figure*}[h] 
  \begin{subfigure}[t]{0.5\textwidth}
    \centering
    \makebox[\linewidth][l]{\textbf{(a)}} \\
    \input{figures/exp3_acc.pgf}
    \phantomcaption 
    \label{ch5:fig:exp3-acc}
\end{subfigure}
  \begin{subfigure}[t]{0.5\textwidth}
    \centering 
    \makebox[\linewidth][l]{\textbf{(b)}} \\
    \input{figures/exp3_mean_rank.pgf}
    \phantomcaption 
    \label{ch5:fig:exp3-mr}
  \end{subfigure}
    \caption[\gls{ar} accuracy (a) and \gls{ar} mean rank (b) at the end of the training for each frequency reliability combination.]{\gls{ar} accuracy (a) and \gls{ar} mean rank (b) at the end of the training for each frequency reliability combination. Scores are represented in colors, with brighter colors representing higher scores.}
\end{figure*}

\section{Effect of Dataset and Tokenizer}\label{appendix:tok-data}
In the main experiments, we used the pretrained GPT2 tokenizer \cite{radford2019language} and the CC100 dataset \cite{conneau-etal-2020-unsupervised}. A natural question is whether or not the reported results hold if the models are trained on a different dataset and/or a dataset tokenized by a different tokenizer. Since replicating the series of experiments with various combinations of datasets and tokenizers is beyond our compute budget, as a proxy experiment, we report frequency \paba{A}{B} and reliability \pbgivenaba{A}{B} of various datasets with different tokenizers. Our datasets include CC100 \cite{conneau-etal-2020-unsupervised}, The Pile \cite{pile2021}, Dolma \cite{soldaini-etal-2024-dolma}, and FineWeb \cite{penedo2024finewebdatasetsdecantingweb}, and our tokenizers include GPT2 \cite{radford2019language}, LLaMA3 \cite{grattafiori2024llama3herdmodels}, and T5 \cite{raffel2023exploringlimitstransferlearning}.

\Cref{fig:effect-dataset} summarizes, for each tokenizer, the effect of the choice of corpus on the 2 metrics: frequency and reliability. Across different tokenizers and sequence lengths, both frequency and reliability largely follow a consistent pattern: Pile > Dolma > FineWeb > CC-100. Although this warrants a more thorough investigation, a quick observation is that the web-crawl corpus (CC-100 and FineWeb) seems to have fewer repetitions than mixed domain corpora (Dolma and Pile). Given that the mean sequence length is Dolma > Pile > CC-100 > FineWeb, we surmise that this could be due to a domain effect rather than a simple document length effect. However, given that the Pile corpus has a much wider spread, and that Dolma and FineWeb are virtually indistinguishable in the 2 metrics, these observations should be taken with caution.

\Cref{fig:effect-tokenizer} similarly summarizes the effect of tokenizer on frequency and reliability for each of the corpora. With the exception of the Pile corpus, across different corpora and sequence lengths, frequency \paba{A}{B} seems to follow a consistent pattern: T5 > GPT2 > LLaMA-3. This is rather intuitive; T5, GPT2, and LLaMA-3 have vocabulary sizes of 32128, 50275, and 128256, respectively. The smaller the vocabulary size, the more likely a given word is tokenized into multiple tokens, resulting in more repetitions in a given sequence. Note that the shaded regions in \Cref{fig:effect-dataset} and \Cref{fig:effect-tokenizer} are based on the standard deviations over the sequences, and with a large enough sample size (> 10M tokens) used in this analysis, the standard error is so small that it would be invisible in the figures. Hence, the differences across corpora and tokenizers seem robust, and further investigation is needed to fully understand their effect on the repetitiveness of the training data, and more importantly, the emergence of \glspl{ih} as well as phase transition in general.
\begin{figure}[t]
\centering
    \input{figures/appendix_effect_dataset.pgf}
    \caption{Effect of the choice of dataset on frequency \paba{A}{B} (top) and reliability \pbgivenaba{A}{B} (bottom). Each column represents the tokenizer used to tokenize the dataset. The shaded region represents the standard deviation across sequences of the size indicated on the $x$-axis.}
    \label{fig:effect-dataset}
\end{figure}
\begin{figure}[t]
\centering
    \input{figures/appendix_effect_tokenizer.pgf}
    \caption{Effect of the choice of tokenizer on frequency \paba{A}{B} (top) and reliability \pbgivenaba{A}{B} (bottom). Each column represents the dataset. The shaded region represents the standard deviation across sequences of the size indicated on the $x$-axis.}
    \label{fig:effect-tokenizer}
\end{figure}

\begin{table*}[t]
    \centering
    \caption{The full statistics of the effect of sequence length, corpus, and tokenizer on frequency and reliability. $P_A$ and $P_B$ refer to \paba{A}{B} and \pbgivenaba{A}{B}, respectively. $\mu$ and $\sigma$ are sample mean and standard deviation across sequences, respectively.}
    {\fontsize{8}{9.5}\selectfont
    \input{tables/corpus_tok}
    \label{tab:corpus-tok}}
\end{table*}


\end{document}

%% file: figures/emergence_var.pgf
\begingroup%
\makeatletter%
\begin{pgfpicture}%
\pgfpathrectangle{\pgfpointorigin}{\pgfqpoint{1.724816in}{1.842063in}}%
\pgfusepath{use as bounding box, clip}%
\begin{pgfscope}%
\pgfsetbuttcap%
\pgfsetmiterjoin%
\definecolor{currentfill}{rgb}{1.000000,1.000000,1.000000}%
\pgfsetfillcolor{currentfill}%
\pgfsetlinewidth{0.000000pt}%
\definecolor{currentstroke}{rgb}{1.000000,1.000000,1.000000}%
\pgfsetstrokecolor{currentstroke}%
\pgfsetdash{}{0pt}%
\pgfpathmoveto{\pgfqpoint{0.000000in}{0.000000in}}%
\pgfpathlineto{\pgfqpoint{1.724816in}{0.000000in}}%
\pgfpathlineto{\pgfqpoint{1.724816in}{1.842063in}}%
\pgfpathlineto{\pgfqpoint{0.000000in}{1.842063in}}%
\pgfpathlineto{\pgfqpoint{0.000000in}{0.000000in}}%
\pgfpathclose%
\pgfusepath{fill}%
\end{pgfscope}%
\begin{pgfscope}%
\pgfsetbuttcap%
\pgfsetmiterjoin%
\definecolor{currentfill}{rgb}{1.000000,1.000000,1.000000}%
\pgfsetfillcolor{currentfill}%
\pgfsetlinewidth{0.000000pt}%
\definecolor{currentstroke}{rgb}{0.000000,0.000000,0.000000}%
\pgfsetstrokecolor{currentstroke}%
\pgfsetstrokeopacity{0.000000}%
\pgfsetdash{}{0pt}%
\pgfpathmoveto{\pgfqpoint{0.539816in}{0.587063in}}%
\pgfpathlineto{\pgfqpoint{1.624816in}{0.587063in}}%
\pgfpathlineto{\pgfqpoint{1.624816in}{1.742063in}}%
\pgfpathlineto{\pgfqpoint{0.539816in}{1.742063in}}%
\pgfpathlineto{\pgfqpoint{0.539816in}{0.587063in}}%
\pgfpathclose%
\pgfusepath{fill}%
\end{pgfscope}%
\begin{pgfscope}%
\pgfpathrectangle{\pgfqpoint{0.539816in}{0.587063in}}{\pgfqpoint{1.085000in}{1.155000in}}%
\pgfusepath{clip}%
\pgfsetbuttcap%
\pgfsetroundjoin%
\definecolor{currentfill}{rgb}{0.121569,0.466667,0.705882}%
\pgfsetfillcolor{currentfill}%
\pgfsetfillopacity{0.700000}%
\pgfsetlinewidth{0.501875pt}%
\definecolor{currentstroke}{rgb}{1.000000,1.000000,1.000000}%
\pgfsetstrokecolor{currentstroke}%
\pgfsetstrokeopacity{0.700000}%
\pgfsetdash{}{0pt}%
\pgfsys@defobject{currentmarker}{\pgfqpoint{-0.029463in}{-0.029463in}}{\pgfqpoint{0.029463in}{0.029463in}}{%
\pgfpathmoveto{\pgfqpoint{0.000000in}{-0.029463in}}%
\pgfpathcurveto{\pgfqpoint{0.007814in}{-0.029463in}}{\pgfqpoint{0.015308in}{-0.026358in}}{\pgfqpoint{0.020833in}{-0.020833in}}%
\pgfpathcurveto{\pgfqpoint{0.026358in}{-0.015308in}}{\pgfqpoint{0.029463in}{-0.007814in}}{\pgfqpoint{0.029463in}{0.000000in}}%
\pgfpathcurveto{\pgfqpoint{0.029463in}{0.007814in}}{\pgfqpoint{0.026358in}{0.015308in}}{\pgfqpoint{0.020833in}{0.020833in}}%
\pgfpathcurveto{\pgfqpoint{0.015308in}{0.026358in}}{\pgfqpoint{0.007814in}{0.029463in}}{\pgfqpoint{0.000000in}{0.029463in}}%
\pgfpathcurveto{\pgfqpoint{-0.007814in}{0.029463in}}{\pgfqpoint{-0.015308in}{0.026358in}}{\pgfqpoint{-0.020833in}{0.020833in}}%
\pgfpathcurveto{\pgfqpoint{-0.026358in}{0.015308in}}{\pgfqpoint{-0.029463in}{0.007814in}}{\pgfqpoint{-0.029463in}{0.000000in}}%
\pgfpathcurveto{\pgfqpoint{-0.029463in}{-0.007814in}}{\pgfqpoint{-0.026358in}{-0.015308in}}{\pgfqpoint{-0.020833in}{-0.020833in}}%
\pgfpathcurveto{\pgfqpoint{-0.015308in}{-0.026358in}}{\pgfqpoint{-0.007814in}{-0.029463in}}{\pgfqpoint{0.000000in}{-0.029463in}}%
\pgfpathlineto{\pgfqpoint{0.000000in}{-0.029463in}}%
\pgfpathclose%
\pgfusepath{stroke,fill}%
}%
\begin{pgfscope}%
\pgfsys@transformshift{0.589134in}{1.654627in}%
\pgfsys@useobject{currentmarker}{}%
\end{pgfscope}%
\begin{pgfscope}%
\pgfsys@transformshift{0.775240in}{1.278809in}%
\pgfsys@useobject{currentmarker}{}%
\end{pgfscope}%
\begin{pgfscope}%
\pgfsys@transformshift{0.961347in}{1.089118in}%
\pgfsys@useobject{currentmarker}{}%
\end{pgfscope}%
\begin{pgfscope}%
\pgfsys@transformshift{1.147453in}{1.048443in}%
\pgfsys@useobject{currentmarker}{}%
\end{pgfscope}%
\begin{pgfscope}%
\pgfsys@transformshift{1.333559in}{0.739367in}%
\pgfsys@useobject{currentmarker}{}%
\end{pgfscope}%
\begin{pgfscope}%
\pgfsys@transformshift{1.519666in}{0.679935in}%
\pgfsys@useobject{currentmarker}{}%
\end{pgfscope}%
\end{pgfscope}%
\begin{pgfscope}%
\pgfpathrectangle{\pgfqpoint{0.539816in}{0.587063in}}{\pgfqpoint{1.085000in}{1.155000in}}%
\pgfusepath{clip}%
\pgfsetbuttcap%
\pgfsetroundjoin%
\definecolor{currentfill}{rgb}{1.000000,0.498039,0.054902}%
\pgfsetfillcolor{currentfill}%
\pgfsetfillopacity{0.700000}%
\pgfsetlinewidth{0.501875pt}%
\definecolor{currentstroke}{rgb}{1.000000,1.000000,1.000000}%
\pgfsetstrokecolor{currentstroke}%
\pgfsetstrokeopacity{0.700000}%
\pgfsetdash{}{0pt}%
\pgfsys@defobject{currentmarker}{\pgfqpoint{-0.029463in}{-0.029463in}}{\pgfqpoint{0.029463in}{0.029463in}}{%
\pgfpathmoveto{\pgfqpoint{0.000000in}{0.029463in}}%
\pgfpathlineto{\pgfqpoint{-0.029463in}{-0.029463in}}%
\pgfpathlineto{\pgfqpoint{0.029463in}{-0.029463in}}%
\pgfpathlineto{\pgfqpoint{0.000000in}{0.029463in}}%
\pgfpathclose%
\pgfusepath{stroke,fill}%
}%
\begin{pgfscope}%
\pgfsys@transformshift{0.617050in}{1.680534in}%
\pgfsys@useobject{currentmarker}{}%
\end{pgfscope}%
\begin{pgfscope}%
\pgfsys@transformshift{0.803156in}{1.360090in}%
\pgfsys@useobject{currentmarker}{}%
\end{pgfscope}%
\begin{pgfscope}%
\pgfsys@transformshift{0.989263in}{1.061755in}%
\pgfsys@useobject{currentmarker}{}%
\end{pgfscope}%
\begin{pgfscope}%
\pgfsys@transformshift{1.175369in}{1.056862in}%
\pgfsys@useobject{currentmarker}{}%
\end{pgfscope}%
\begin{pgfscope}%
\pgfsys@transformshift{1.361475in}{0.713858in}%
\pgfsys@useobject{currentmarker}{}%
\end{pgfscope}%
\begin{pgfscope}%
\pgfsys@transformshift{1.547582in}{0.639563in}%
\pgfsys@useobject{currentmarker}{}%
\end{pgfscope}%
\end{pgfscope}%
\begin{pgfscope}%
\pgfpathrectangle{\pgfqpoint{0.539816in}{0.587063in}}{\pgfqpoint{1.085000in}{1.155000in}}%
\pgfusepath{clip}%
\pgfsetbuttcap%
\pgfsetroundjoin%
\definecolor{currentfill}{rgb}{0.172549,0.627451,0.172549}%
\pgfsetfillcolor{currentfill}%
\pgfsetfillopacity{0.700000}%
\pgfsetlinewidth{0.501875pt}%
\definecolor{currentstroke}{rgb}{1.000000,1.000000,1.000000}%
\pgfsetstrokecolor{currentstroke}%
\pgfsetstrokeopacity{0.700000}%
\pgfsetdash{}{0pt}%
\pgfsys@defobject{currentmarker}{\pgfqpoint{-0.029463in}{-0.029463in}}{\pgfqpoint{0.029463in}{0.029463in}}{%
\pgfpathmoveto{\pgfqpoint{-0.029463in}{-0.029463in}}%
\pgfpathlineto{\pgfqpoint{0.029463in}{-0.029463in}}%
\pgfpathlineto{\pgfqpoint{0.029463in}{0.029463in}}%
\pgfpathlineto{\pgfqpoint{-0.029463in}{0.029463in}}%
\pgfpathlineto{\pgfqpoint{-0.029463in}{-0.029463in}}%
\pgfpathclose%
\pgfusepath{stroke,fill}%
}%
\begin{pgfscope}%
\pgfsys@transformshift{0.644966in}{1.598094in}%
\pgfsys@useobject{currentmarker}{}%
\end{pgfscope}%
\begin{pgfscope}%
\pgfsys@transformshift{0.831072in}{1.288099in}%
\pgfsys@useobject{currentmarker}{}%
\end{pgfscope}%
\begin{pgfscope}%
\pgfsys@transformshift{1.017178in}{1.102118in}%
\pgfsys@useobject{currentmarker}{}%
\end{pgfscope}%
\begin{pgfscope}%
\pgfsys@transformshift{1.203285in}{1.046228in}%
\pgfsys@useobject{currentmarker}{}%
\end{pgfscope}%
\begin{pgfscope}%
\pgfsys@transformshift{1.389391in}{0.710374in}%
\pgfsys@useobject{currentmarker}{}%
\end{pgfscope}%
\begin{pgfscope}%
\pgfsys@transformshift{1.575498in}{0.681915in}%
\pgfsys@useobject{currentmarker}{}%
\end{pgfscope}%
\end{pgfscope}%
\begin{pgfscope}%
\pgfpathrectangle{\pgfqpoint{0.539816in}{0.587063in}}{\pgfqpoint{1.085000in}{1.155000in}}%
\pgfusepath{clip}%
\pgfsetrectcap%
\pgfsetroundjoin%
\pgfsetlinewidth{0.803000pt}%
\definecolor{currentstroke}{rgb}{0.690196,0.690196,0.690196}%
\pgfsetstrokecolor{currentstroke}%
\pgfsetstrokeopacity{0.250000}%
\pgfsetdash{}{0pt}%
\pgfpathmoveto{\pgfqpoint{0.617050in}{0.587063in}}%
\pgfpathlineto{\pgfqpoint{0.617050in}{1.742063in}}%
\pgfusepath{stroke}%
\end{pgfscope}%
\begin{pgfscope}%
\pgfsetbuttcap%
\pgfsetroundjoin%
\definecolor{currentfill}{rgb}{0.000000,0.000000,0.000000}%
\pgfsetfillcolor{currentfill}%
\pgfsetlinewidth{0.803000pt}%
\definecolor{currentstroke}{rgb}{0.000000,0.000000,0.000000}%
\pgfsetstrokecolor{currentstroke}%
\pgfsetdash{}{0pt}%
\pgfsys@defobject{currentmarker}{\pgfqpoint{0.000000in}{-0.048611in}}{\pgfqpoint{0.000000in}{0.000000in}}{%
\pgfpathmoveto{\pgfqpoint{0.000000in}{0.000000in}}%
\pgfpathlineto{\pgfqpoint{0.000000in}{-0.048611in}}%
\pgfusepath{stroke,fill}%
}%
\begin{pgfscope}%
\pgfsys@transformshift{0.617050in}{0.587063in}%
\pgfsys@useobject{currentmarker}{}%
\end{pgfscope}%
\end{pgfscope}%
\begin{pgfscope}%
\definecolor{textcolor}{rgb}{0.000000,0.000000,0.000000}%
\pgfsetstrokecolor{textcolor}%
\pgfsetfillcolor{textcolor}%
\pgftext[x=0.617050in,y=0.489841in,right,top,rotate=45.000000]{\color{textcolor}{\rmfamily\fontsize{8.000000}{9.600000}\selectfont\catcode`\^=\active\def^{\ifmmode\sp\else\^{}\fi}\catcode`\%=\active\def
\end{pgfscope}%
\begin{pgfscope}%
\pgfpathrectangle{\pgfqpoint{0.539816in}{0.587063in}}{\pgfqpoint{1.085000in}{1.155000in}}%
\pgfusepath{clip}%
\pgfsetrectcap%
\pgfsetroundjoin%
\pgfsetlinewidth{0.803000pt}%
\definecolor{currentstroke}{rgb}{0.690196,0.690196,0.690196}%
\pgfsetstrokecolor{currentstroke}%
\pgfsetstrokeopacity{0.250000}%
\pgfsetdash{}{0pt}%
\pgfpathmoveto{\pgfqpoint{0.803156in}{0.587063in}}%
\pgfpathlineto{\pgfqpoint{0.803156in}{1.742063in}}%
\pgfusepath{stroke}%
\end{pgfscope}%
\begin{pgfscope}%
\pgfsetbuttcap%
\pgfsetroundjoin%
\definecolor{currentfill}{rgb}{0.000000,0.000000,0.000000}%
\pgfsetfillcolor{currentfill}%
\pgfsetlinewidth{0.803000pt}%
\definecolor{currentstroke}{rgb}{0.000000,0.000000,0.000000}%
\pgfsetstrokecolor{currentstroke}%
\pgfsetdash{}{0pt}%
\pgfsys@defobject{currentmarker}{\pgfqpoint{0.000000in}{-0.048611in}}{\pgfqpoint{0.000000in}{0.000000in}}{%
\pgfpathmoveto{\pgfqpoint{0.000000in}{0.000000in}}%
\pgfpathlineto{\pgfqpoint{0.000000in}{-0.048611in}}%
\pgfusepath{stroke,fill}%
}%
\begin{pgfscope}%
\pgfsys@transformshift{0.803156in}{0.587063in}%
\pgfsys@useobject{currentmarker}{}%
\end{pgfscope}%
\end{pgfscope}%
\begin{pgfscope}%
\definecolor{textcolor}{rgb}{0.000000,0.000000,0.000000}%
\pgfsetstrokecolor{textcolor}%
\pgfsetfillcolor{textcolor}%
\pgftext[x=0.803156in,y=0.489841in,right,top,rotate=45.000000]{\color{textcolor}{\rmfamily\fontsize{8.000000}{9.600000}\selectfont\catcode`\^=\active\def^{\ifmmode\sp\else\^{}\fi}\catcode`\%=\active\def
\end{pgfscope}%
\begin{pgfscope}%
\pgfpathrectangle{\pgfqpoint{0.539816in}{0.587063in}}{\pgfqpoint{1.085000in}{1.155000in}}%
\pgfusepath{clip}%
\pgfsetrectcap%
\pgfsetroundjoin%
\pgfsetlinewidth{0.803000pt}%
\definecolor{currentstroke}{rgb}{0.690196,0.690196,0.690196}%
\pgfsetstrokecolor{currentstroke}%
\pgfsetstrokeopacity{0.250000}%
\pgfsetdash{}{0pt}%
\pgfpathmoveto{\pgfqpoint{0.989263in}{0.587063in}}%
\pgfpathlineto{\pgfqpoint{0.989263in}{1.742063in}}%
\pgfusepath{stroke}%
\end{pgfscope}%
\begin{pgfscope}%
\pgfsetbuttcap%
\pgfsetroundjoin%
\definecolor{currentfill}{rgb}{0.000000,0.000000,0.000000}%
\pgfsetfillcolor{currentfill}%
\pgfsetlinewidth{0.803000pt}%
\definecolor{currentstroke}{rgb}{0.000000,0.000000,0.000000}%
\pgfsetstrokecolor{currentstroke}%
\pgfsetdash{}{0pt}%
\pgfsys@defobject{currentmarker}{\pgfqpoint{0.000000in}{-0.048611in}}{\pgfqpoint{0.000000in}{0.000000in}}{%
\pgfpathmoveto{\pgfqpoint{0.000000in}{0.000000in}}%
\pgfpathlineto{\pgfqpoint{0.000000in}{-0.048611in}}%
\pgfusepath{stroke,fill}%
}%
\begin{pgfscope}%
\pgfsys@transformshift{0.989263in}{0.587063in}%
\pgfsys@useobject{currentmarker}{}%
\end{pgfscope}%
\end{pgfscope}%
\begin{pgfscope}%
\definecolor{textcolor}{rgb}{0.000000,0.000000,0.000000}%
\pgfsetstrokecolor{textcolor}%
\pgfsetfillcolor{textcolor}%
\pgftext[x=0.989263in,y=0.489841in,right,top,rotate=45.000000]{\color{textcolor}{\rmfamily\fontsize{8.000000}{9.600000}\selectfont\catcode`\^=\active\def^{\ifmmode\sp\else\^{}\fi}\catcode`\%=\active\def
\end{pgfscope}%
\begin{pgfscope}%
\pgfpathrectangle{\pgfqpoint{0.539816in}{0.587063in}}{\pgfqpoint{1.085000in}{1.155000in}}%
\pgfusepath{clip}%
\pgfsetrectcap%
\pgfsetroundjoin%
\pgfsetlinewidth{0.803000pt}%
\definecolor{currentstroke}{rgb}{0.690196,0.690196,0.690196}%
\pgfsetstrokecolor{currentstroke}%
\pgfsetstrokeopacity{0.250000}%
\pgfsetdash{}{0pt}%
\pgfpathmoveto{\pgfqpoint{1.175369in}{0.587063in}}%
\pgfpathlineto{\pgfqpoint{1.175369in}{1.742063in}}%
\pgfusepath{stroke}%
\end{pgfscope}%
\begin{pgfscope}%
\pgfsetbuttcap%
\pgfsetroundjoin%
\definecolor{currentfill}{rgb}{0.000000,0.000000,0.000000}%
\pgfsetfillcolor{currentfill}%
\pgfsetlinewidth{0.803000pt}%
\definecolor{currentstroke}{rgb}{0.000000,0.000000,0.000000}%
\pgfsetstrokecolor{currentstroke}%
\pgfsetdash{}{0pt}%
\pgfsys@defobject{currentmarker}{\pgfqpoint{0.000000in}{-0.048611in}}{\pgfqpoint{0.000000in}{0.000000in}}{%
\pgfpathmoveto{\pgfqpoint{0.000000in}{0.000000in}}%
\pgfpathlineto{\pgfqpoint{0.000000in}{-0.048611in}}%
\pgfusepath{stroke,fill}%
}%
\begin{pgfscope}%
\pgfsys@transformshift{1.175369in}{0.587063in}%
\pgfsys@useobject{currentmarker}{}%
\end{pgfscope}%
\end{pgfscope}%
\begin{pgfscope}%
\definecolor{textcolor}{rgb}{0.000000,0.000000,0.000000}%
\pgfsetstrokecolor{textcolor}%
\pgfsetfillcolor{textcolor}%
\pgftext[x=1.175369in,y=0.489841in,right,top,rotate=45.000000]{\color{textcolor}{\rmfamily\fontsize{8.000000}{9.600000}\selectfont\catcode`\^=\active\def^{\ifmmode\sp\else\^{}\fi}\catcode`\%=\active\def
\end{pgfscope}%
\begin{pgfscope}%
\pgfpathrectangle{\pgfqpoint{0.539816in}{0.587063in}}{\pgfqpoint{1.085000in}{1.155000in}}%
\pgfusepath{clip}%
\pgfsetrectcap%
\pgfsetroundjoin%
\pgfsetlinewidth{0.803000pt}%
\definecolor{currentstroke}{rgb}{0.690196,0.690196,0.690196}%
\pgfsetstrokecolor{currentstroke}%
\pgfsetstrokeopacity{0.250000}%
\pgfsetdash{}{0pt}%
\pgfpathmoveto{\pgfqpoint{1.361475in}{0.587063in}}%
\pgfpathlineto{\pgfqpoint{1.361475in}{1.742063in}}%
\pgfusepath{stroke}%
\end{pgfscope}%
\begin{pgfscope}%
\pgfsetbuttcap%
\pgfsetroundjoin%
\definecolor{currentfill}{rgb}{0.000000,0.000000,0.000000}%
\pgfsetfillcolor{currentfill}%
\pgfsetlinewidth{0.803000pt}%
\definecolor{currentstroke}{rgb}{0.000000,0.000000,0.000000}%
\pgfsetstrokecolor{currentstroke}%
\pgfsetdash{}{0pt}%
\pgfsys@defobject{currentmarker}{\pgfqpoint{0.000000in}{-0.048611in}}{\pgfqpoint{0.000000in}{0.000000in}}{%
\pgfpathmoveto{\pgfqpoint{0.000000in}{0.000000in}}%
\pgfpathlineto{\pgfqpoint{0.000000in}{-0.048611in}}%
\pgfusepath{stroke,fill}%
}%
\begin{pgfscope}%
\pgfsys@transformshift{1.361475in}{0.587063in}%
\pgfsys@useobject{currentmarker}{}%
\end{pgfscope}%
\end{pgfscope}%
\begin{pgfscope}%
\definecolor{textcolor}{rgb}{0.000000,0.000000,0.000000}%
\pgfsetstrokecolor{textcolor}%
\pgfsetfillcolor{textcolor}%
\pgftext[x=1.361475in,y=0.489841in,right,top,rotate=45.000000]{\color{textcolor}{\rmfamily\fontsize{8.000000}{9.600000}\selectfont\catcode`\^=\active\def^{\ifmmode\sp\else\^{}\fi}\catcode`\%=\active\def
\end{pgfscope}%
\begin{pgfscope}%
\pgfpathrectangle{\pgfqpoint{0.539816in}{0.587063in}}{\pgfqpoint{1.085000in}{1.155000in}}%
\pgfusepath{clip}%
\pgfsetrectcap%
\pgfsetroundjoin%
\pgfsetlinewidth{0.803000pt}%
\definecolor{currentstroke}{rgb}{0.690196,0.690196,0.690196}%
\pgfsetstrokecolor{currentstroke}%
\pgfsetstrokeopacity{0.250000}%
\pgfsetdash{}{0pt}%
\pgfpathmoveto{\pgfqpoint{1.547582in}{0.587063in}}%
\pgfpathlineto{\pgfqpoint{1.547582in}{1.742063in}}%
\pgfusepath{stroke}%
\end{pgfscope}%
\begin{pgfscope}%
\pgfsetbuttcap%
\pgfsetroundjoin%
\definecolor{currentfill}{rgb}{0.000000,0.000000,0.000000}%
\pgfsetfillcolor{currentfill}%
\pgfsetlinewidth{0.803000pt}%
\definecolor{currentstroke}{rgb}{0.000000,0.000000,0.000000}%
\pgfsetstrokecolor{currentstroke}%
\pgfsetdash{}{0pt}%
\pgfsys@defobject{currentmarker}{\pgfqpoint{0.000000in}{-0.048611in}}{\pgfqpoint{0.000000in}{0.000000in}}{%
\pgfpathmoveto{\pgfqpoint{0.000000in}{0.000000in}}%
\pgfpathlineto{\pgfqpoint{0.000000in}{-0.048611in}}%
\pgfusepath{stroke,fill}%
}%
\begin{pgfscope}%
\pgfsys@transformshift{1.547582in}{0.587063in}%
\pgfsys@useobject{currentmarker}{}%
\end{pgfscope}%
\end{pgfscope}%
\begin{pgfscope}%
\definecolor{textcolor}{rgb}{0.000000,0.000000,0.000000}%
\pgfsetstrokecolor{textcolor}%
\pgfsetfillcolor{textcolor}%
\pgftext[x=1.547582in,y=0.489841in,right,top,rotate=45.000000]{\color{textcolor}{\rmfamily\fontsize{8.000000}{9.600000}\selectfont\catcode`\^=\active\def^{\ifmmode\sp\else\^{}\fi}\catcode`\%=\active\def
\end{pgfscope}%
\begin{pgfscope}%
\pgfpathrectangle{\pgfqpoint{0.539816in}{0.587063in}}{\pgfqpoint{1.085000in}{1.155000in}}%
\pgfusepath{clip}%
\pgfsetrectcap%
\pgfsetroundjoin%
\pgfsetlinewidth{0.803000pt}%
\definecolor{currentstroke}{rgb}{0.690196,0.690196,0.690196}%
\pgfsetstrokecolor{currentstroke}%
\pgfsetstrokeopacity{0.250000}%
\pgfsetdash{}{0pt}%
\pgfpathmoveto{\pgfqpoint{0.539816in}{0.729996in}}%
\pgfpathlineto{\pgfqpoint{1.624816in}{0.729996in}}%
\pgfusepath{stroke}%
\end{pgfscope}%
\begin{pgfscope}%
\pgfsetbuttcap%
\pgfsetroundjoin%
\definecolor{currentfill}{rgb}{0.000000,0.000000,0.000000}%
\pgfsetfillcolor{currentfill}%
\pgfsetlinewidth{0.803000pt}%
\definecolor{currentstroke}{rgb}{0.000000,0.000000,0.000000}%
\pgfsetstrokecolor{currentstroke}%
\pgfsetdash{}{0pt}%
\pgfsys@defobject{currentmarker}{\pgfqpoint{-0.048611in}{0.000000in}}{\pgfqpoint{-0.000000in}{0.000000in}}{%
\pgfpathmoveto{\pgfqpoint{-0.000000in}{0.000000in}}%
\pgfpathlineto{\pgfqpoint{-0.048611in}{0.000000in}}%
\pgfusepath{stroke,fill}%
}%
\begin{pgfscope}%
\pgfsys@transformshift{0.539816in}{0.729996in}%
\pgfsys@useobject{currentmarker}{}%
\end{pgfscope}%
\end{pgfscope}%
\begin{pgfscope}%
\definecolor{textcolor}{rgb}{0.000000,0.000000,0.000000}%
\pgfsetstrokecolor{textcolor}%
\pgfsetfillcolor{textcolor}%
\pgftext[x=0.266667in, y=0.690868in, left, base]{\color{textcolor}{\rmfamily\fontsize{8.000000}{9.600000}\selectfont\catcode`\^=\active\def^{\ifmmode\sp\else\^{}\fi}\catcode`\%=\active\def
\end{pgfscope}%
\begin{pgfscope}%
\pgfpathrectangle{\pgfqpoint{0.539816in}{0.587063in}}{\pgfqpoint{1.085000in}{1.155000in}}%
\pgfusepath{clip}%
\pgfsetrectcap%
\pgfsetroundjoin%
\pgfsetlinewidth{0.803000pt}%
\definecolor{currentstroke}{rgb}{0.690196,0.690196,0.690196}%
\pgfsetstrokecolor{currentstroke}%
\pgfsetstrokeopacity{0.250000}%
\pgfsetdash{}{0pt}%
\pgfpathmoveto{\pgfqpoint{0.539816in}{1.429311in}}%
\pgfpathlineto{\pgfqpoint{1.624816in}{1.429311in}}%
\pgfusepath{stroke}%
\end{pgfscope}%
\begin{pgfscope}%
\pgfsetbuttcap%
\pgfsetroundjoin%
\definecolor{currentfill}{rgb}{0.000000,0.000000,0.000000}%
\pgfsetfillcolor{currentfill}%
\pgfsetlinewidth{0.803000pt}%
\definecolor{currentstroke}{rgb}{0.000000,0.000000,0.000000}%
\pgfsetstrokecolor{currentstroke}%
\pgfsetdash{}{0pt}%
\pgfsys@defobject{currentmarker}{\pgfqpoint{-0.048611in}{0.000000in}}{\pgfqpoint{-0.000000in}{0.000000in}}{%
\pgfpathmoveto{\pgfqpoint{-0.000000in}{0.000000in}}%
\pgfpathlineto{\pgfqpoint{-0.048611in}{0.000000in}}%
\pgfusepath{stroke,fill}%
}%
\begin{pgfscope}%
\pgfsys@transformshift{0.539816in}{1.429311in}%
\pgfsys@useobject{currentmarker}{}%
\end{pgfscope}%
\end{pgfscope}%
\begin{pgfscope}%
\definecolor{textcolor}{rgb}{0.000000,0.000000,0.000000}%
\pgfsetstrokecolor{textcolor}%
\pgfsetfillcolor{textcolor}%
\pgftext[x=0.266667in, y=1.390183in, left, base]{\color{textcolor}{\rmfamily\fontsize{8.000000}{9.600000}\selectfont\catcode`\^=\active\def^{\ifmmode\sp\else\^{}\fi}\catcode`\%=\active\def
\end{pgfscope}%
\begin{pgfscope}%
\pgfpathrectangle{\pgfqpoint{0.539816in}{0.587063in}}{\pgfqpoint{1.085000in}{1.155000in}}%
\pgfusepath{clip}%
\pgfsetrectcap%
\pgfsetroundjoin%
\pgfsetlinewidth{0.803000pt}%
\definecolor{currentstroke}{rgb}{0.690196,0.690196,0.690196}%
\pgfsetstrokecolor{currentstroke}%
\pgfsetstrokeopacity{0.250000}%
\pgfsetdash{}{0pt}%
\pgfpathmoveto{\pgfqpoint{0.539816in}{0.621670in}}%
\pgfpathlineto{\pgfqpoint{1.624816in}{0.621670in}}%
\pgfusepath{stroke}%
\end{pgfscope}%
\begin{pgfscope}%
\pgfsetbuttcap%
\pgfsetroundjoin%
\definecolor{currentfill}{rgb}{0.000000,0.000000,0.000000}%
\pgfsetfillcolor{currentfill}%
\pgfsetlinewidth{0.602250pt}%
\definecolor{currentstroke}{rgb}{0.000000,0.000000,0.000000}%
\pgfsetstrokecolor{currentstroke}%
\pgfsetdash{}{0pt}%
\pgfsys@defobject{currentmarker}{\pgfqpoint{-0.027778in}{0.000000in}}{\pgfqpoint{-0.000000in}{0.000000in}}{%
\pgfpathmoveto{\pgfqpoint{-0.000000in}{0.000000in}}%
\pgfpathlineto{\pgfqpoint{-0.027778in}{0.000000in}}%
\pgfusepath{stroke,fill}%
}%
\begin{pgfscope}%
\pgfsys@transformshift{0.539816in}{0.621670in}%
\pgfsys@useobject{currentmarker}{}%
\end{pgfscope}%
\end{pgfscope}%
\begin{pgfscope}%
\pgfpathrectangle{\pgfqpoint{0.539816in}{0.587063in}}{\pgfqpoint{1.085000in}{1.155000in}}%
\pgfusepath{clip}%
\pgfsetrectcap%
\pgfsetroundjoin%
\pgfsetlinewidth{0.803000pt}%
\definecolor{currentstroke}{rgb}{0.690196,0.690196,0.690196}%
\pgfsetstrokecolor{currentstroke}%
\pgfsetstrokeopacity{0.250000}%
\pgfsetdash{}{0pt}%
\pgfpathmoveto{\pgfqpoint{0.539816in}{0.662225in}}%
\pgfpathlineto{\pgfqpoint{1.624816in}{0.662225in}}%
\pgfusepath{stroke}%
\end{pgfscope}%
\begin{pgfscope}%
\pgfsetbuttcap%
\pgfsetroundjoin%
\definecolor{currentfill}{rgb}{0.000000,0.000000,0.000000}%
\pgfsetfillcolor{currentfill}%
\pgfsetlinewidth{0.602250pt}%
\definecolor{currentstroke}{rgb}{0.000000,0.000000,0.000000}%
\pgfsetstrokecolor{currentstroke}%
\pgfsetdash{}{0pt}%
\pgfsys@defobject{currentmarker}{\pgfqpoint{-0.027778in}{0.000000in}}{\pgfqpoint{-0.000000in}{0.000000in}}{%
\pgfpathmoveto{\pgfqpoint{-0.000000in}{0.000000in}}%
\pgfpathlineto{\pgfqpoint{-0.027778in}{0.000000in}}%
\pgfusepath{stroke,fill}%
}%
\begin{pgfscope}%
\pgfsys@transformshift{0.539816in}{0.662225in}%
\pgfsys@useobject{currentmarker}{}%
\end{pgfscope}%
\end{pgfscope}%
\begin{pgfscope}%
\pgfpathrectangle{\pgfqpoint{0.539816in}{0.587063in}}{\pgfqpoint{1.085000in}{1.155000in}}%
\pgfusepath{clip}%
\pgfsetrectcap%
\pgfsetroundjoin%
\pgfsetlinewidth{0.803000pt}%
\definecolor{currentstroke}{rgb}{0.690196,0.690196,0.690196}%
\pgfsetstrokecolor{currentstroke}%
\pgfsetstrokeopacity{0.250000}%
\pgfsetdash{}{0pt}%
\pgfpathmoveto{\pgfqpoint{0.539816in}{0.697997in}}%
\pgfpathlineto{\pgfqpoint{1.624816in}{0.697997in}}%
\pgfusepath{stroke}%
\end{pgfscope}%
\begin{pgfscope}%
\pgfsetbuttcap%
\pgfsetroundjoin%
\definecolor{currentfill}{rgb}{0.000000,0.000000,0.000000}%
\pgfsetfillcolor{currentfill}%
\pgfsetlinewidth{0.602250pt}%
\definecolor{currentstroke}{rgb}{0.000000,0.000000,0.000000}%
\pgfsetstrokecolor{currentstroke}%
\pgfsetdash{}{0pt}%
\pgfsys@defobject{currentmarker}{\pgfqpoint{-0.027778in}{0.000000in}}{\pgfqpoint{-0.000000in}{0.000000in}}{%
\pgfpathmoveto{\pgfqpoint{-0.000000in}{0.000000in}}%
\pgfpathlineto{\pgfqpoint{-0.027778in}{0.000000in}}%
\pgfusepath{stroke,fill}%
}%
\begin{pgfscope}%
\pgfsys@transformshift{0.539816in}{0.697997in}%
\pgfsys@useobject{currentmarker}{}%
\end{pgfscope}%
\end{pgfscope}%
\begin{pgfscope}%
\pgfpathrectangle{\pgfqpoint{0.539816in}{0.587063in}}{\pgfqpoint{1.085000in}{1.155000in}}%
\pgfusepath{clip}%
\pgfsetrectcap%
\pgfsetroundjoin%
\pgfsetlinewidth{0.803000pt}%
\definecolor{currentstroke}{rgb}{0.690196,0.690196,0.690196}%
\pgfsetstrokecolor{currentstroke}%
\pgfsetstrokeopacity{0.250000}%
\pgfsetdash{}{0pt}%
\pgfpathmoveto{\pgfqpoint{0.539816in}{0.940511in}}%
\pgfpathlineto{\pgfqpoint{1.624816in}{0.940511in}}%
\pgfusepath{stroke}%
\end{pgfscope}%
\begin{pgfscope}%
\pgfsetbuttcap%
\pgfsetroundjoin%
\definecolor{currentfill}{rgb}{0.000000,0.000000,0.000000}%
\pgfsetfillcolor{currentfill}%
\pgfsetlinewidth{0.602250pt}%
\definecolor{currentstroke}{rgb}{0.000000,0.000000,0.000000}%
\pgfsetstrokecolor{currentstroke}%
\pgfsetdash{}{0pt}%
\pgfsys@defobject{currentmarker}{\pgfqpoint{-0.027778in}{0.000000in}}{\pgfqpoint{-0.000000in}{0.000000in}}{%
\pgfpathmoveto{\pgfqpoint{-0.000000in}{0.000000in}}%
\pgfpathlineto{\pgfqpoint{-0.027778in}{0.000000in}}%
\pgfusepath{stroke,fill}%
}%
\begin{pgfscope}%
\pgfsys@transformshift{0.539816in}{0.940511in}%
\pgfsys@useobject{currentmarker}{}%
\end{pgfscope}%
\end{pgfscope}%
\begin{pgfscope}%
\pgfpathrectangle{\pgfqpoint{0.539816in}{0.587063in}}{\pgfqpoint{1.085000in}{1.155000in}}%
\pgfusepath{clip}%
\pgfsetrectcap%
\pgfsetroundjoin%
\pgfsetlinewidth{0.803000pt}%
\definecolor{currentstroke}{rgb}{0.690196,0.690196,0.690196}%
\pgfsetstrokecolor{currentstroke}%
\pgfsetstrokeopacity{0.250000}%
\pgfsetdash{}{0pt}%
\pgfpathmoveto{\pgfqpoint{0.539816in}{1.063654in}}%
\pgfpathlineto{\pgfqpoint{1.624816in}{1.063654in}}%
\pgfusepath{stroke}%
\end{pgfscope}%
\begin{pgfscope}%
\pgfsetbuttcap%
\pgfsetroundjoin%
\definecolor{currentfill}{rgb}{0.000000,0.000000,0.000000}%
\pgfsetfillcolor{currentfill}%
\pgfsetlinewidth{0.602250pt}%
\definecolor{currentstroke}{rgb}{0.000000,0.000000,0.000000}%
\pgfsetstrokecolor{currentstroke}%
\pgfsetdash{}{0pt}%
\pgfsys@defobject{currentmarker}{\pgfqpoint{-0.027778in}{0.000000in}}{\pgfqpoint{-0.000000in}{0.000000in}}{%
\pgfpathmoveto{\pgfqpoint{-0.000000in}{0.000000in}}%
\pgfpathlineto{\pgfqpoint{-0.027778in}{0.000000in}}%
\pgfusepath{stroke,fill}%
}%
\begin{pgfscope}%
\pgfsys@transformshift{0.539816in}{1.063654in}%
\pgfsys@useobject{currentmarker}{}%
\end{pgfscope}%
\end{pgfscope}%
\begin{pgfscope}%
\pgfpathrectangle{\pgfqpoint{0.539816in}{0.587063in}}{\pgfqpoint{1.085000in}{1.155000in}}%
\pgfusepath{clip}%
\pgfsetrectcap%
\pgfsetroundjoin%
\pgfsetlinewidth{0.803000pt}%
\definecolor{currentstroke}{rgb}{0.690196,0.690196,0.690196}%
\pgfsetstrokecolor{currentstroke}%
\pgfsetstrokeopacity{0.250000}%
\pgfsetdash{}{0pt}%
\pgfpathmoveto{\pgfqpoint{0.539816in}{1.151025in}}%
\pgfpathlineto{\pgfqpoint{1.624816in}{1.151025in}}%
\pgfusepath{stroke}%
\end{pgfscope}%
\begin{pgfscope}%
\pgfsetbuttcap%
\pgfsetroundjoin%
\definecolor{currentfill}{rgb}{0.000000,0.000000,0.000000}%
\pgfsetfillcolor{currentfill}%
\pgfsetlinewidth{0.602250pt}%
\definecolor{currentstroke}{rgb}{0.000000,0.000000,0.000000}%
\pgfsetstrokecolor{currentstroke}%
\pgfsetdash{}{0pt}%
\pgfsys@defobject{currentmarker}{\pgfqpoint{-0.027778in}{0.000000in}}{\pgfqpoint{-0.000000in}{0.000000in}}{%
\pgfpathmoveto{\pgfqpoint{-0.000000in}{0.000000in}}%
\pgfpathlineto{\pgfqpoint{-0.027778in}{0.000000in}}%
\pgfusepath{stroke,fill}%
}%
\begin{pgfscope}%
\pgfsys@transformshift{0.539816in}{1.151025in}%
\pgfsys@useobject{currentmarker}{}%
\end{pgfscope}%
\end{pgfscope}%
\begin{pgfscope}%
\pgfpathrectangle{\pgfqpoint{0.539816in}{0.587063in}}{\pgfqpoint{1.085000in}{1.155000in}}%
\pgfusepath{clip}%
\pgfsetrectcap%
\pgfsetroundjoin%
\pgfsetlinewidth{0.803000pt}%
\definecolor{currentstroke}{rgb}{0.690196,0.690196,0.690196}%
\pgfsetstrokecolor{currentstroke}%
\pgfsetstrokeopacity{0.250000}%
\pgfsetdash{}{0pt}%
\pgfpathmoveto{\pgfqpoint{0.539816in}{1.218796in}}%
\pgfpathlineto{\pgfqpoint{1.624816in}{1.218796in}}%
\pgfusepath{stroke}%
\end{pgfscope}%
\begin{pgfscope}%
\pgfsetbuttcap%
\pgfsetroundjoin%
\definecolor{currentfill}{rgb}{0.000000,0.000000,0.000000}%
\pgfsetfillcolor{currentfill}%
\pgfsetlinewidth{0.602250pt}%
\definecolor{currentstroke}{rgb}{0.000000,0.000000,0.000000}%
\pgfsetstrokecolor{currentstroke}%
\pgfsetdash{}{0pt}%
\pgfsys@defobject{currentmarker}{\pgfqpoint{-0.027778in}{0.000000in}}{\pgfqpoint{-0.000000in}{0.000000in}}{%
\pgfpathmoveto{\pgfqpoint{-0.000000in}{0.000000in}}%
\pgfpathlineto{\pgfqpoint{-0.027778in}{0.000000in}}%
\pgfusepath{stroke,fill}%
}%
\begin{pgfscope}%
\pgfsys@transformshift{0.539816in}{1.218796in}%
\pgfsys@useobject{currentmarker}{}%
\end{pgfscope}%
\end{pgfscope}%
\begin{pgfscope}%
\pgfpathrectangle{\pgfqpoint{0.539816in}{0.587063in}}{\pgfqpoint{1.085000in}{1.155000in}}%
\pgfusepath{clip}%
\pgfsetrectcap%
\pgfsetroundjoin%
\pgfsetlinewidth{0.803000pt}%
\definecolor{currentstroke}{rgb}{0.690196,0.690196,0.690196}%
\pgfsetstrokecolor{currentstroke}%
\pgfsetstrokeopacity{0.250000}%
\pgfsetdash{}{0pt}%
\pgfpathmoveto{\pgfqpoint{0.539816in}{1.274169in}}%
\pgfpathlineto{\pgfqpoint{1.624816in}{1.274169in}}%
\pgfusepath{stroke}%
\end{pgfscope}%
\begin{pgfscope}%
\pgfsetbuttcap%
\pgfsetroundjoin%
\definecolor{currentfill}{rgb}{0.000000,0.000000,0.000000}%
\pgfsetfillcolor{currentfill}%
\pgfsetlinewidth{0.602250pt}%
\definecolor{currentstroke}{rgb}{0.000000,0.000000,0.000000}%
\pgfsetstrokecolor{currentstroke}%
\pgfsetdash{}{0pt}%
\pgfsys@defobject{currentmarker}{\pgfqpoint{-0.027778in}{0.000000in}}{\pgfqpoint{-0.000000in}{0.000000in}}{%
\pgfpathmoveto{\pgfqpoint{-0.000000in}{0.000000in}}%
\pgfpathlineto{\pgfqpoint{-0.027778in}{0.000000in}}%
\pgfusepath{stroke,fill}%
}%
\begin{pgfscope}%
\pgfsys@transformshift{0.539816in}{1.274169in}%
\pgfsys@useobject{currentmarker}{}%
\end{pgfscope}%
\end{pgfscope}%
\begin{pgfscope}%
\pgfpathrectangle{\pgfqpoint{0.539816in}{0.587063in}}{\pgfqpoint{1.085000in}{1.155000in}}%
\pgfusepath{clip}%
\pgfsetrectcap%
\pgfsetroundjoin%
\pgfsetlinewidth{0.803000pt}%
\definecolor{currentstroke}{rgb}{0.690196,0.690196,0.690196}%
\pgfsetstrokecolor{currentstroke}%
\pgfsetstrokeopacity{0.250000}%
\pgfsetdash{}{0pt}%
\pgfpathmoveto{\pgfqpoint{0.539816in}{1.320986in}}%
\pgfpathlineto{\pgfqpoint{1.624816in}{1.320986in}}%
\pgfusepath{stroke}%
\end{pgfscope}%
\begin{pgfscope}%
\pgfsetbuttcap%
\pgfsetroundjoin%
\definecolor{currentfill}{rgb}{0.000000,0.000000,0.000000}%
\pgfsetfillcolor{currentfill}%
\pgfsetlinewidth{0.602250pt}%
\definecolor{currentstroke}{rgb}{0.000000,0.000000,0.000000}%
\pgfsetstrokecolor{currentstroke}%
\pgfsetdash{}{0pt}%
\pgfsys@defobject{currentmarker}{\pgfqpoint{-0.027778in}{0.000000in}}{\pgfqpoint{-0.000000in}{0.000000in}}{%
\pgfpathmoveto{\pgfqpoint{-0.000000in}{0.000000in}}%
\pgfpathlineto{\pgfqpoint{-0.027778in}{0.000000in}}%
\pgfusepath{stroke,fill}%
}%
\begin{pgfscope}%
\pgfsys@transformshift{0.539816in}{1.320986in}%
\pgfsys@useobject{currentmarker}{}%
\end{pgfscope}%
\end{pgfscope}%
\begin{pgfscope}%
\pgfpathrectangle{\pgfqpoint{0.539816in}{0.587063in}}{\pgfqpoint{1.085000in}{1.155000in}}%
\pgfusepath{clip}%
\pgfsetrectcap%
\pgfsetroundjoin%
\pgfsetlinewidth{0.803000pt}%
\definecolor{currentstroke}{rgb}{0.690196,0.690196,0.690196}%
\pgfsetstrokecolor{currentstroke}%
\pgfsetstrokeopacity{0.250000}%
\pgfsetdash{}{0pt}%
\pgfpathmoveto{\pgfqpoint{0.539816in}{1.361540in}}%
\pgfpathlineto{\pgfqpoint{1.624816in}{1.361540in}}%
\pgfusepath{stroke}%
\end{pgfscope}%
\begin{pgfscope}%
\pgfsetbuttcap%
\pgfsetroundjoin%
\definecolor{currentfill}{rgb}{0.000000,0.000000,0.000000}%
\pgfsetfillcolor{currentfill}%
\pgfsetlinewidth{0.602250pt}%
\definecolor{currentstroke}{rgb}{0.000000,0.000000,0.000000}%
\pgfsetstrokecolor{currentstroke}%
\pgfsetdash{}{0pt}%
\pgfsys@defobject{currentmarker}{\pgfqpoint{-0.027778in}{0.000000in}}{\pgfqpoint{-0.000000in}{0.000000in}}{%
\pgfpathmoveto{\pgfqpoint{-0.000000in}{0.000000in}}%
\pgfpathlineto{\pgfqpoint{-0.027778in}{0.000000in}}%
\pgfusepath{stroke,fill}%
}%
\begin{pgfscope}%
\pgfsys@transformshift{0.539816in}{1.361540in}%
\pgfsys@useobject{currentmarker}{}%
\end{pgfscope}%
\end{pgfscope}%
\begin{pgfscope}%
\pgfpathrectangle{\pgfqpoint{0.539816in}{0.587063in}}{\pgfqpoint{1.085000in}{1.155000in}}%
\pgfusepath{clip}%
\pgfsetrectcap%
\pgfsetroundjoin%
\pgfsetlinewidth{0.803000pt}%
\definecolor{currentstroke}{rgb}{0.690196,0.690196,0.690196}%
\pgfsetstrokecolor{currentstroke}%
\pgfsetstrokeopacity{0.250000}%
\pgfsetdash{}{0pt}%
\pgfpathmoveto{\pgfqpoint{0.539816in}{1.397312in}}%
\pgfpathlineto{\pgfqpoint{1.624816in}{1.397312in}}%
\pgfusepath{stroke}%
\end{pgfscope}%
\begin{pgfscope}%
\pgfsetbuttcap%
\pgfsetroundjoin%
\definecolor{currentfill}{rgb}{0.000000,0.000000,0.000000}%
\pgfsetfillcolor{currentfill}%
\pgfsetlinewidth{0.602250pt}%
\definecolor{currentstroke}{rgb}{0.000000,0.000000,0.000000}%
\pgfsetstrokecolor{currentstroke}%
\pgfsetdash{}{0pt}%
\pgfsys@defobject{currentmarker}{\pgfqpoint{-0.027778in}{0.000000in}}{\pgfqpoint{-0.000000in}{0.000000in}}{%
\pgfpathmoveto{\pgfqpoint{-0.000000in}{0.000000in}}%
\pgfpathlineto{\pgfqpoint{-0.027778in}{0.000000in}}%
\pgfusepath{stroke,fill}%
}%
\begin{pgfscope}%
\pgfsys@transformshift{0.539816in}{1.397312in}%
\pgfsys@useobject{currentmarker}{}%
\end{pgfscope}%
\end{pgfscope}%
\begin{pgfscope}%
\pgfpathrectangle{\pgfqpoint{0.539816in}{0.587063in}}{\pgfqpoint{1.085000in}{1.155000in}}%
\pgfusepath{clip}%
\pgfsetrectcap%
\pgfsetroundjoin%
\pgfsetlinewidth{0.803000pt}%
\definecolor{currentstroke}{rgb}{0.690196,0.690196,0.690196}%
\pgfsetstrokecolor{currentstroke}%
\pgfsetstrokeopacity{0.250000}%
\pgfsetdash{}{0pt}%
\pgfpathmoveto{\pgfqpoint{0.539816in}{1.639826in}}%
\pgfpathlineto{\pgfqpoint{1.624816in}{1.639826in}}%
\pgfusepath{stroke}%
\end{pgfscope}%
\begin{pgfscope}%
\pgfsetbuttcap%
\pgfsetroundjoin%
\definecolor{currentfill}{rgb}{0.000000,0.000000,0.000000}%
\pgfsetfillcolor{currentfill}%
\pgfsetlinewidth{0.602250pt}%
\definecolor{currentstroke}{rgb}{0.000000,0.000000,0.000000}%
\pgfsetstrokecolor{currentstroke}%
\pgfsetdash{}{0pt}%
\pgfsys@defobject{currentmarker}{\pgfqpoint{-0.027778in}{0.000000in}}{\pgfqpoint{-0.000000in}{0.000000in}}{%
\pgfpathmoveto{\pgfqpoint{-0.000000in}{0.000000in}}%
\pgfpathlineto{\pgfqpoint{-0.027778in}{0.000000in}}%
\pgfusepath{stroke,fill}%
}%
\begin{pgfscope}%
\pgfsys@transformshift{0.539816in}{1.639826in}%
\pgfsys@useobject{currentmarker}{}%
\end{pgfscope}%
\end{pgfscope}%
\begin{pgfscope}%
\definecolor{textcolor}{rgb}{0.000000,0.000000,0.000000}%
\pgfsetstrokecolor{textcolor}%
\pgfsetfillcolor{textcolor}%
\pgftext[x=0.211111in,y=1.164563in,,bottom,rotate=90.000000]{\color{textcolor}{\rmfamily\fontsize{9.000000}{10.800000}\selectfont\catcode`\^=\active\def^{\ifmmode\sp\else\^{}\fi}\catcode`\%=\active\def
\end{pgfscope}%
\begin{pgfscope}%
\pgfsetrectcap%
\pgfsetmiterjoin%
\pgfsetlinewidth{0.803000pt}%
\definecolor{currentstroke}{rgb}{0.000000,0.000000,0.000000}%
\pgfsetstrokecolor{currentstroke}%
\pgfsetdash{}{0pt}%
\pgfpathmoveto{\pgfqpoint{0.539816in}{0.587063in}}%
\pgfpathlineto{\pgfqpoint{0.539816in}{1.742063in}}%
\pgfusepath{stroke}%
\end{pgfscope}%
\begin{pgfscope}%
\pgfsetrectcap%
\pgfsetmiterjoin%
\pgfsetlinewidth{0.803000pt}%
\definecolor{currentstroke}{rgb}{0.000000,0.000000,0.000000}%
\pgfsetstrokecolor{currentstroke}%
\pgfsetdash{}{0pt}%
\pgfpathmoveto{\pgfqpoint{1.624816in}{0.587063in}}%
\pgfpathlineto{\pgfqpoint{1.624816in}{1.742063in}}%
\pgfusepath{stroke}%
\end{pgfscope}%
\begin{pgfscope}%
\pgfsetrectcap%
\pgfsetmiterjoin%
\pgfsetlinewidth{0.803000pt}%
\definecolor{currentstroke}{rgb}{0.000000,0.000000,0.000000}%
\pgfsetstrokecolor{currentstroke}%
\pgfsetdash{}{0pt}%
\pgfpathmoveto{\pgfqpoint{0.539816in}{0.587063in}}%
\pgfpathlineto{\pgfqpoint{1.624816in}{0.587063in}}%
\pgfusepath{stroke}%
\end{pgfscope}%
\begin{pgfscope}%
\pgfsetrectcap%
\pgfsetmiterjoin%
\pgfsetlinewidth{0.803000pt}%
\definecolor{currentstroke}{rgb}{0.000000,0.000000,0.000000}%
\pgfsetstrokecolor{currentstroke}%
\pgfsetdash{}{0pt}%
\pgfpathmoveto{\pgfqpoint{0.539816in}{1.742063in}}%
\pgfpathlineto{\pgfqpoint{1.624816in}{1.742063in}}%
\pgfusepath{stroke}%
\end{pgfscope}%
\begin{pgfscope}%
\pgfpathrectangle{\pgfqpoint{0.539816in}{0.587063in}}{\pgfqpoint{1.085000in}{1.155000in}}%
\pgfusepath{clip}%
\pgfsetbuttcap%
\pgfsetroundjoin%
\pgfsetlinewidth{1.204500pt}%
\definecolor{currentstroke}{rgb}{0.411765,0.411765,0.411765}%
\pgfsetstrokecolor{currentstroke}%
\pgfsetdash{}{0pt}%
\pgfpathmoveto{\pgfqpoint{0.617050in}{1.595918in}}%
\pgfpathlineto{\pgfqpoint{0.617050in}{1.689563in}}%
\pgfusepath{stroke}%
\end{pgfscope}%
\begin{pgfscope}%
\pgfpathrectangle{\pgfqpoint{0.539816in}{0.587063in}}{\pgfqpoint{1.085000in}{1.155000in}}%
\pgfusepath{clip}%
\pgfsetbuttcap%
\pgfsetroundjoin%
\pgfsetlinewidth{1.204500pt}%
\definecolor{currentstroke}{rgb}{0.411765,0.411765,0.411765}%
\pgfsetstrokecolor{currentstroke}%
\pgfsetdash{}{0pt}%
\pgfpathmoveto{\pgfqpoint{0.803156in}{1.253979in}}%
\pgfpathlineto{\pgfqpoint{0.803156in}{1.359375in}}%
\pgfusepath{stroke}%
\end{pgfscope}%
\begin{pgfscope}%
\pgfpathrectangle{\pgfqpoint{0.539816in}{0.587063in}}{\pgfqpoint{1.085000in}{1.155000in}}%
\pgfusepath{clip}%
\pgfsetbuttcap%
\pgfsetroundjoin%
\pgfsetlinewidth{1.204500pt}%
\definecolor{currentstroke}{rgb}{0.411765,0.411765,0.411765}%
\pgfsetstrokecolor{currentstroke}%
\pgfsetdash{}{0pt}%
\pgfpathmoveto{\pgfqpoint{0.989263in}{1.060832in}}%
\pgfpathlineto{\pgfqpoint{0.989263in}{1.107000in}}%
\pgfusepath{stroke}%
\end{pgfscope}%
\begin{pgfscope}%
\pgfpathrectangle{\pgfqpoint{0.539816in}{0.587063in}}{\pgfqpoint{1.085000in}{1.155000in}}%
\pgfusepath{clip}%
\pgfsetbuttcap%
\pgfsetroundjoin%
\pgfsetlinewidth{1.204500pt}%
\definecolor{currentstroke}{rgb}{0.411765,0.411765,0.411765}%
\pgfsetstrokecolor{currentstroke}%
\pgfsetdash{}{0pt}%
\pgfpathmoveto{\pgfqpoint{1.175369in}{1.044101in}}%
\pgfpathlineto{\pgfqpoint{1.175369in}{1.056856in}}%
\pgfusepath{stroke}%
\end{pgfscope}%
\begin{pgfscope}%
\pgfpathrectangle{\pgfqpoint{0.539816in}{0.587063in}}{\pgfqpoint{1.085000in}{1.155000in}}%
\pgfusepath{clip}%
\pgfsetbuttcap%
\pgfsetroundjoin%
\pgfsetlinewidth{1.204500pt}%
\definecolor{currentstroke}{rgb}{0.411765,0.411765,0.411765}%
\pgfsetstrokecolor{currentstroke}%
\pgfsetdash{}{0pt}%
\pgfpathmoveto{\pgfqpoint{1.361475in}{0.702752in}}%
\pgfpathlineto{\pgfqpoint{1.361475in}{0.739115in}}%
\pgfusepath{stroke}%
\end{pgfscope}%
\begin{pgfscope}%
\pgfpathrectangle{\pgfqpoint{0.539816in}{0.587063in}}{\pgfqpoint{1.085000in}{1.155000in}}%
\pgfusepath{clip}%
\pgfsetbuttcap%
\pgfsetroundjoin%
\pgfsetlinewidth{1.204500pt}%
\definecolor{currentstroke}{rgb}{0.411765,0.411765,0.411765}%
\pgfsetstrokecolor{currentstroke}%
\pgfsetdash{}{0pt}%
\pgfpathmoveto{\pgfqpoint{1.547582in}{0.640125in}}%
\pgfpathlineto{\pgfqpoint{1.547582in}{0.693079in}}%
\pgfusepath{stroke}%
\end{pgfscope}%
\begin{pgfscope}%
\pgfpathrectangle{\pgfqpoint{0.539816in}{0.587063in}}{\pgfqpoint{1.085000in}{1.155000in}}%
\pgfusepath{clip}%
\pgfsetbuttcap%
\pgfsetroundjoin%
\definecolor{currentfill}{rgb}{0.411765,0.411765,0.411765}%
\pgfsetfillcolor{currentfill}%
\pgfsetlinewidth{1.003750pt}%
\definecolor{currentstroke}{rgb}{0.411765,0.411765,0.411765}%
\pgfsetstrokecolor{currentstroke}%
\pgfsetdash{}{0pt}%
\pgfsys@defobject{currentmarker}{\pgfqpoint{-0.069444in}{-0.000000in}}{\pgfqpoint{0.069444in}{0.000000in}}{%
\pgfpathmoveto{\pgfqpoint{0.069444in}{-0.000000in}}%
\pgfpathlineto{\pgfqpoint{-0.069444in}{0.000000in}}%
\pgfusepath{stroke,fill}%
}%
\begin{pgfscope}%
\pgfsys@transformshift{0.617050in}{1.595918in}%
\pgfsys@useobject{currentmarker}{}%
\end{pgfscope}%
\begin{pgfscope}%
\pgfsys@transformshift{0.803156in}{1.253979in}%
\pgfsys@useobject{currentmarker}{}%
\end{pgfscope}%
\begin{pgfscope}%
\pgfsys@transformshift{0.989263in}{1.060832in}%
\pgfsys@useobject{currentmarker}{}%
\end{pgfscope}%
\begin{pgfscope}%
\pgfsys@transformshift{1.175369in}{1.044101in}%
\pgfsys@useobject{currentmarker}{}%
\end{pgfscope}%
\begin{pgfscope}%
\pgfsys@transformshift{1.361475in}{0.702752in}%
\pgfsys@useobject{currentmarker}{}%
\end{pgfscope}%
\begin{pgfscope}%
\pgfsys@transformshift{1.547582in}{0.640125in}%
\pgfsys@useobject{currentmarker}{}%
\end{pgfscope}%
\end{pgfscope}%
\begin{pgfscope}%
\pgfpathrectangle{\pgfqpoint{0.539816in}{0.587063in}}{\pgfqpoint{1.085000in}{1.155000in}}%
\pgfusepath{clip}%
\pgfsetbuttcap%
\pgfsetroundjoin%
\definecolor{currentfill}{rgb}{0.411765,0.411765,0.411765}%
\pgfsetfillcolor{currentfill}%
\pgfsetlinewidth{1.003750pt}%
\definecolor{currentstroke}{rgb}{0.411765,0.411765,0.411765}%
\pgfsetstrokecolor{currentstroke}%
\pgfsetdash{}{0pt}%
\pgfsys@defobject{currentmarker}{\pgfqpoint{-0.069444in}{-0.000000in}}{\pgfqpoint{0.069444in}{0.000000in}}{%
\pgfpathmoveto{\pgfqpoint{0.069444in}{-0.000000in}}%
\pgfpathlineto{\pgfqpoint{-0.069444in}{0.000000in}}%
\pgfusepath{stroke,fill}%
}%
\begin{pgfscope}%
\pgfsys@transformshift{0.617050in}{1.689563in}%
\pgfsys@useobject{currentmarker}{}%
\end{pgfscope}%
\begin{pgfscope}%
\pgfsys@transformshift{0.803156in}{1.359375in}%
\pgfsys@useobject{currentmarker}{}%
\end{pgfscope}%
\begin{pgfscope}%
\pgfsys@transformshift{0.989263in}{1.107000in}%
\pgfsys@useobject{currentmarker}{}%
\end{pgfscope}%
\begin{pgfscope}%
\pgfsys@transformshift{1.175369in}{1.056856in}%
\pgfsys@useobject{currentmarker}{}%
\end{pgfscope}%
\begin{pgfscope}%
\pgfsys@transformshift{1.361475in}{0.739115in}%
\pgfsys@useobject{currentmarker}{}%
\end{pgfscope}%
\begin{pgfscope}%
\pgfsys@transformshift{1.547582in}{0.693079in}%
\pgfsys@useobject{currentmarker}{}%
\end{pgfscope}%
\end{pgfscope}%
\begin{pgfscope}%
\pgfsetbuttcap%
\pgfsetmiterjoin%
\definecolor{currentfill}{rgb}{1.000000,1.000000,1.000000}%
\pgfsetfillcolor{currentfill}%
\pgfsetfillopacity{0.800000}%
\pgfsetlinewidth{1.003750pt}%
\definecolor{currentstroke}{rgb}{0.800000,0.800000,0.800000}%
\pgfsetstrokecolor{currentstroke}%
\pgfsetstrokeopacity{0.800000}%
\pgfsetdash{}{0pt}%
\pgfpathmoveto{\pgfqpoint{1.052134in}{1.376785in}}%
\pgfpathlineto{\pgfqpoint{1.566482in}{1.376785in}}%
\pgfpathquadraticcurveto{\pgfqpoint{1.583149in}{1.376785in}}{\pgfqpoint{1.583149in}{1.393452in}}%
\pgfpathlineto{\pgfqpoint{1.583149in}{1.683730in}}%
\pgfpathquadraticcurveto{\pgfqpoint{1.583149in}{1.700396in}}{\pgfqpoint{1.566482in}{1.700396in}}%
\pgfpathlineto{\pgfqpoint{1.052134in}{1.700396in}}%
\pgfpathquadraticcurveto{\pgfqpoint{1.035467in}{1.700396in}}{\pgfqpoint{1.035467in}{1.683730in}}%
\pgfpathlineto{\pgfqpoint{1.035467in}{1.393452in}}%
\pgfpathquadraticcurveto{\pgfqpoint{1.035467in}{1.376785in}}{\pgfqpoint{1.052134in}{1.376785in}}%
\pgfpathlineto{\pgfqpoint{1.052134in}{1.376785in}}%
\pgfpathclose%
\pgfusepath{stroke,fill}%
\end{pgfscope}%
\begin{pgfscope}%
\pgfsetbuttcap%
\pgfsetroundjoin%
\definecolor{currentfill}{rgb}{0.121569,0.466667,0.705882}%
\pgfsetfillcolor{currentfill}%
\pgfsetlinewidth{0.501875pt}%
\definecolor{currentstroke}{rgb}{1.000000,1.000000,1.000000}%
\pgfsetstrokecolor{currentstroke}%
\pgfsetdash{}{0pt}%
\pgfsys@defobject{currentmarker}{\pgfqpoint{-0.034021in}{-0.034021in}}{\pgfqpoint{0.034021in}{0.034021in}}{%
\pgfpathmoveto{\pgfqpoint{0.000000in}{-0.034021in}}%
\pgfpathcurveto{\pgfqpoint{0.009022in}{-0.034021in}}{\pgfqpoint{0.017676in}{-0.030436in}}{\pgfqpoint{0.024056in}{-0.024056in}}%
\pgfpathcurveto{\pgfqpoint{0.030436in}{-0.017676in}}{\pgfqpoint{0.034021in}{-0.009022in}}{\pgfqpoint{0.034021in}{0.000000in}}%
\pgfpathcurveto{\pgfqpoint{0.034021in}{0.009022in}}{\pgfqpoint{0.030436in}{0.017676in}}{\pgfqpoint{0.024056in}{0.024056in}}%
\pgfpathcurveto{\pgfqpoint{0.017676in}{0.030436in}}{\pgfqpoint{0.009022in}{0.034021in}}{\pgfqpoint{0.000000in}{0.034021in}}%
\pgfpathcurveto{\pgfqpoint{-0.009022in}{0.034021in}}{\pgfqpoint{-0.017676in}{0.030436in}}{\pgfqpoint{-0.024056in}{0.024056in}}%
\pgfpathcurveto{\pgfqpoint{-0.030436in}{0.017676in}}{\pgfqpoint{-0.034021in}{0.009022in}}{\pgfqpoint{-0.034021in}{0.000000in}}%
\pgfpathcurveto{\pgfqpoint{-0.034021in}{-0.009022in}}{\pgfqpoint{-0.030436in}{-0.017676in}}{\pgfqpoint{-0.024056in}{-0.024056in}}%
\pgfpathcurveto{\pgfqpoint{-0.017676in}{-0.030436in}}{\pgfqpoint{-0.009022in}{-0.034021in}}{\pgfqpoint{0.000000in}{-0.034021in}}%
\pgfpathlineto{\pgfqpoint{0.000000in}{-0.034021in}}%
\pgfpathclose%
\pgfusepath{stroke,fill}%
}%
\begin{pgfscope}%
\pgfsys@transformshift{1.131301in}{1.630605in}%
\pgfsys@useobject{currentmarker}{}%
\end{pgfscope}%
\end{pgfscope}%
\begin{pgfscope}%
\definecolor{textcolor}{rgb}{0.000000,0.000000,0.000000}%
\pgfsetstrokecolor{textcolor}%
\pgfsetfillcolor{textcolor}%
\pgftext[x=1.260468in,y=1.608730in,left,base]{\color{textcolor}{\rmfamily\fontsize{6.000000}{7.200000}\selectfont\catcode`\^=\active\def^{\ifmmode\sp\else\^{}\fi}\catcode`\%=\active\def
\end{pgfscope}%
\begin{pgfscope}%
\pgfsetbuttcap%
\pgfsetroundjoin%
\definecolor{currentfill}{rgb}{1.000000,0.498039,0.054902}%
\pgfsetfillcolor{currentfill}%
\pgfsetlinewidth{0.501875pt}%
\definecolor{currentstroke}{rgb}{1.000000,1.000000,1.000000}%
\pgfsetstrokecolor{currentstroke}%
\pgfsetdash{}{0pt}%
\pgfsys@defobject{currentmarker}{\pgfqpoint{-0.034021in}{-0.034021in}}{\pgfqpoint{0.034021in}{0.034021in}}{%
\pgfpathmoveto{\pgfqpoint{0.000000in}{0.034021in}}%
\pgfpathlineto{\pgfqpoint{-0.034021in}{-0.034021in}}%
\pgfpathlineto{\pgfqpoint{0.034021in}{-0.034021in}}%
\pgfpathlineto{\pgfqpoint{0.000000in}{0.034021in}}%
\pgfpathclose%
\pgfusepath{stroke,fill}%
}%
\begin{pgfscope}%
\pgfsys@transformshift{1.131301in}{1.539401in}%
\pgfsys@useobject{currentmarker}{}%
\end{pgfscope}%
\end{pgfscope}%
\begin{pgfscope}%
\definecolor{textcolor}{rgb}{0.000000,0.000000,0.000000}%
\pgfsetstrokecolor{textcolor}%
\pgfsetfillcolor{textcolor}%
\pgftext[x=1.260468in,y=1.517526in,left,base]{\color{textcolor}{\rmfamily\fontsize{6.000000}{7.200000}\selectfont\catcode`\^=\active\def^{\ifmmode\sp\else\^{}\fi}\catcode`\%=\active\def
\end{pgfscope}%
\begin{pgfscope}%
\pgfsetbuttcap%
\pgfsetroundjoin%
\definecolor{currentfill}{rgb}{0.172549,0.627451,0.172549}%
\pgfsetfillcolor{currentfill}%
\pgfsetlinewidth{0.501875pt}%
\definecolor{currentstroke}{rgb}{1.000000,1.000000,1.000000}%
\pgfsetstrokecolor{currentstroke}%
\pgfsetdash{}{0pt}%
\pgfsys@defobject{currentmarker}{\pgfqpoint{-0.034021in}{-0.034021in}}{\pgfqpoint{0.034021in}{0.034021in}}{%
\pgfpathmoveto{\pgfqpoint{-0.034021in}{-0.034021in}}%
\pgfpathlineto{\pgfqpoint{0.034021in}{-0.034021in}}%
\pgfpathlineto{\pgfqpoint{0.034021in}{0.034021in}}%
\pgfpathlineto{\pgfqpoint{-0.034021in}{0.034021in}}%
\pgfpathlineto{\pgfqpoint{-0.034021in}{-0.034021in}}%
\pgfpathclose%
\pgfusepath{stroke,fill}%
}%
\begin{pgfscope}%
\pgfsys@transformshift{1.131301in}{1.448197in}%
\pgfsys@useobject{currentmarker}{}%
\end{pgfscope}%
\end{pgfscope}%
\begin{pgfscope}%
\definecolor{textcolor}{rgb}{0.000000,0.000000,0.000000}%
\pgfsetstrokecolor{textcolor}%
\pgfsetfillcolor{textcolor}%
\pgftext[x=1.260468in,y=1.426322in,left,base]{\color{textcolor}{\rmfamily\fontsize{6.000000}{7.200000}\selectfont\catcode`\^=\active\def^{\ifmmode\sp\else\^{}\fi}\catcode`\%=\active\def
\end{pgfscope}%
\end{pgfpicture}%
\makeatother%
\endgroup%

%% file: tables/dists.tex
\begin{tabular}{r|lcc|lrrrl}
\toprule
     &\multicolumn{3}{c|}{\textbf{Properties}}&\multicolumn{4}{c}{\textbf{Statistics}}  \\
     &Marginal&LD&CAT&H($\cdot$)&Intra-group&Inter-group&$\text{D}_{\text{{KL}}}(\cdot\mid\mid\text{target})$\\
\midrule

\config{Zipf}{+}{+}&\multirow{3}{*}{Zipfian}&\greenchk&\greenchk&6.2142&0.3987&0.1007&0.0001\\
\config{Zipf}{+}{-}&&\greenchk&\redx&6.1988&0.0999&0.1010&0.0001  \\
\config{Zipf}{-}{-}&&\redx&\redx&9.5239&1&1&-  \\
\midrule
\config{Unif}{+}{+}&\multirow{3}{*}{Uniform}&\greenchk&\greenchk&6.5388&0.3719&0.0010&4e-6  \\
\config{Unif}{+}{-}&&\greenchk&\redx&6.4759&0.0873&0.0104&0.0001  \\
\config{Unif}{-}{-}&&\redx&\redx&13.2734&1&1&-  \\
\midrule
\config{Gaus}{+}{+}&\multirow{3}{*}{Gaussian}&\greenchk&\greenchk&6.2435&0.3995&0.1003&0.0001  \\
\config{Gaus}{+}{-}&&\greenchk&\redx&6.2407&0.0999&0.1002&0.0001  \\
\config{Gaus}{-}{-}&&\redx&\redx&12.3490&1&1&-  \\
\bottomrule
\end{tabular}

%% file: tables/hyperparameters.tex
\begin{tabular}{l|rrr}
\toprule
\multicolumn{4}{c}{Architecture}\\
\midrule
\texttt{name}           & GPT2 50M & GPT2 125M & GPT2 350M \\ 
\texttt{vocab\_size}    & 50,257   & 50,257    & 50,257    \\ 
\texttt{context\_size}  & 4--2048    & 32--2048     & 32--2048     \\
\texttt{d\_embed}       & 768      & 1,024     & 1,024     \\
\texttt{d\_ffn}         & 3,072    & 4,096     & 4,096     \\
\texttt{n\_layer}       & 2        & 12        & 24        \\
\texttt{n\_head}        & 8        & 12        & 16        \\
\texttt{activation}     & gelu     & gelu      & gelu      \\ 
\texttt{num\_params}    & 50M      & 125M       & 350M       \\ 
\midrule
\multicolumn{4}{c}{Training}\\
\midrule
\texttt{train\_size}      & 1B & 1B & 1B     \\
\texttt{num\_epoch}       & 1 & 1 & 1      \\ 
\texttt{train\_amount}    & 1B & 1B & 1B     \\
\texttt{batch\_size}      & 4--512  & 16--256 & 16--256 \\
\texttt{weight\_decay}    & 0.1 & 0.1 & 0.1    \\
\texttt{warmup\_steps}    & 1\% & 1\% & 1\%    \\
\texttt{lr}               & 5e-4 & 5e-4& 5e-4  \\
\texttt{lr\_scheduler}    & cosine & cosine & cosine \\
\bottomrule
\end{tabular}

%% file: tables/full_regression.tex
\begin{tabular}{llcccccc}
\toprule
\textbf{y}& \textbf{x} & \textbf{coef} & \textbf{std err} & \textbf{t} & \textbf{\textit{p}} & \multicolumn{2}{c}{\textbf{95\% CI}}  \\
\midrule
\multirow{4}{*}{\textbf{PS}} & \textbf{Intercept} &      \textbf{\textcolor{mdgreen}{13.2700}}  &        0.468     &    \textbf{\textcolor{mdgreen}{28.326}}  &         0.000        &       \textbf{\textcolor{mdgreen}{12.315}}    &       \textbf{\textcolor{mdgreen}{14.225}}     \\
& \textbf{log(B)}      &      \textbf{\textcolor{mdred}{-0.3683}}  &        0.044     &    \textbf{\textcolor{mdred}{-8.318}}  &         0.000        &       \textbf{\textcolor{mdred}{-0.459}}    &       \textbf{\textcolor{mdred}{-0.278}}     \\
& \textbf{log(C)}      &      \textbf{\textcolor{mdred}{-0.6149}}  &        0.073     &    \textbf{\textcolor{mdred}{-8.411}}  &         0.000        &       \textbf{\textcolor{mdred}{-0.764}}    &       \textbf{\textcolor{mdred}{-0.466}}     \\
& \textbf{log(M)}      &      -0.0027  &        0.063     &    -0.043  &         0.966        &       -0.130    &        0.125     \\

\midrule
\multirow{4}{*}{\textbf{LA}} & \textbf{Intercept} &      \textbf{\textcolor{mdgreen}{13.6083}}  &        0.477     &    \textbf{\textcolor{mdgreen}{28.529}}  &         0.000        &       \textbf{\textcolor{mdgreen}{12.635}}    &       \textbf{\textcolor{mdgreen}{14.581}}     \\
& \textbf{log(B)}      &      \textbf{\textcolor{mdred}{-0.3735}}  &        0.045     &    \textbf{\textcolor{mdred}{-8.286}}  &         0.000        &       \textbf{\textcolor{mdred}{-0.465}}    &       \textbf{\textcolor{mdred}{-0.282}}     \\
& \textbf{log(C)}      &      \textbf{\textcolor{mdred}{-0.6461}}  &        0.074     &    \textbf{\textcolor{mdred}{-8.678}}  &         0.000        &       \textbf{\textcolor{mdred}{-0.798}}    &       \textbf{\textcolor{mdred}{-0.494}}     \\
& \textbf{log(M)}      &      -0.0069  &        0.064     &    -0.108  &         0.915        &       -0.137    &        0.123     \\

\midrule
\multirow{4}{*}{\textbf{AR (accuracy)}} & \textbf{Intercept} &      \textbf{\textcolor{mdgreen}{16.3236}}  &        0.632     &    \textbf{\textcolor{mdgreen}{25.846}}  &         0.000        &       \textbf{\textcolor{mdgreen}{15.036}}    &       \textbf{\textcolor{mdgreen}{17.612}}     \\
& \textbf{log(B)}      &      \textbf{\textcolor{mdred}{-0.5885}}  &        0.060     &    \textbf{\textcolor{mdred}{-9.859}}  &         0.000        &       \textbf{\textcolor{mdred}{-0.710}}    &       \textbf{\textcolor{mdred}{-0.467}}     \\
& \textbf{log(C)}      &      \textbf{\textcolor{mdred}{-0.7925}}  &        0.099     &    \textbf{\textcolor{mdred}{-8.040}}  &         0.000        &       \textbf{\textcolor{mdred}{-0.993}}    &       \textbf{\textcolor{mdred}{-0.591}}     \\
& \textbf{log(M)}      &      -0.0154  &        0.084     &    -0.182  &         0.856        &       -0.188    &        0.157     \\

\midrule
\multirow{4}{*}{\textbf{AR (mean rank)}} & \textbf{Intercept} &      \textbf{\textcolor{mdgreen}{12.3780}}  &        0.723     &    \textbf{\textcolor{mdgreen}{17.125}}  &         0.000        &       \textbf{\textcolor{mdgreen}{10.904}}    &       \textbf{\textcolor{mdgreen}{13.852}}     \\
& \textbf{log(B)}      &      \textbf{\textcolor{mdred}{-0.5194}}  &        0.068     &    \textbf{\textcolor{mdred}{-7.603}}  &         0.000        &       \textbf{\textcolor{mdred}{-0.659}}    &       \textbf{\textcolor{mdred}{-0.380}}     \\
& \textbf{log(C)}      &      \textbf{\textcolor{mdred}{-0.4928}}  &        0.113     &    \textbf{\textcolor{mdred}{-4.369}}  &         0.000        &       \textbf{\textcolor{mdred}{-0.723}}    &       \textbf{\textcolor{mdred}{-0.263}}     \\
& \textbf{log(M)}      &       0.1295  &        0.097     &     1.340  &         0.190        &       -0.068    &        0.327     \\

\bottomrule
\end{tabular}

%% file: tables/nomodelsize_regression.tex
\begin{tabular}{llcccccc}
\toprule
                   \textbf{y} & \textbf{x}& \textbf{coef} & \textbf{std err} & \textbf{t} & \textbf{\textit{p}} & \multicolumn{2}{c}{\textbf{95\% CI}}  \\
\midrule
\multirow{3}{*}{\textbf{PS}} & \textbf{Intercept} &      \textbf{\textcolor{mdgreen}{13.2630}}  &        0.432     &    \textbf{\textcolor{mdgreen}{30.666}}  &         0.000        &       \textbf{\textcolor{mdgreen}{12.382}}    &       \textbf{\textcolor{mdgreen}{14.144}}     \\
& \textbf{logB}      &      \textbf{\textcolor{mdred}{-0.3689}}  &        0.041     &    \textbf{\textcolor{mdred}{-9.010}}  &         0.000        &       \textbf{\textcolor{mdred}{-0.452}}    &       \textbf{\textcolor{mdred}{-0.286}}     \\
& \textbf{logC}      &      \textbf{\textcolor{mdred}{-0.6155}}  &        0.071     &    \textbf{\textcolor{mdred}{-8.694}}  &         0.000        &       \textbf{\textcolor{mdred}{-0.760}}    &       \textbf{\textcolor{mdred}{-0.471}}     \\

\midrule
\multirow{3}{*}{\textbf{LA}} & \textbf{Intercept} &      \textbf{\textcolor{mdgreen}{13.5904}}  &        0.440     &    \textbf{\textcolor{mdgreen}{30.856}}  &         0.000        &       \textbf{\textcolor{mdgreen}{12.693}}    &       \textbf{\textcolor{mdgreen}{14.488}}     \\
& \textbf{logB}      &      \textbf{\textcolor{mdred}{-0.3752}}  &        0.042     &    \textbf{\textcolor{mdred}{-8.999}}  &         0.000        &       \textbf{\textcolor{mdred}{-0.460}}    &       \textbf{\textcolor{mdred}{-0.290}}     \\
& \textbf{logC}      &      \textbf{\textcolor{mdred}{-0.6475}}  &        0.072     &    \textbf{\textcolor{mdred}{-8.981}}  &         0.000        &       \textbf{\textcolor{mdred}{-0.794}}    &       \textbf{\textcolor{mdred}{-0.501}}     \\

\midrule
\multirow{3}{*}{\textbf{AR (accuracy)}} & \textbf{Intercept} &      \textbf{\textcolor{mdgreen}{16.2837}}  &        0.583     &    \textbf{\textcolor{mdgreen}{27.913}}  &         0.000        &       \textbf{\textcolor{mdgreen}{15.095}}    &       \textbf{\textcolor{mdgreen}{17.472}}     \\
& \textbf{logB}      &      \textbf{\textcolor{mdred}{-0.5922}}  &        0.055     &   \textbf{\textcolor{mdred}{-10.723}}  &         0.000        &       \textbf{\textcolor{mdred}{-0.705}}    &       \textbf{\textcolor{mdred}{-0.480}}     \\
& \textbf{logC}      &      \textbf{\textcolor{mdred}{-0.7957}}  &        0.095     &    \textbf{\textcolor{mdred}{-8.333}}  &         0.000        &       \textbf{\textcolor{mdred}{-0.990}}    &       \textbf{\textcolor{mdred}{-0.601}}     \\

\midrule
\multirow{3}{*}{\textbf{AR (mean rank)}} & \textbf{Intercept} &      \textbf{\textcolor{mdgreen}{12.7138}}  &        0.686     &    \textbf{\textcolor{mdgreen}{18.525}}  &         0.000        &       \textbf{\textcolor{mdgreen}{11.316}}    &       \textbf{\textcolor{mdgreen}{14.112}}     \\
& \textbf{logB}      &      \textbf{\textcolor{mdred}{-0.4880}}  &        0.065     &    \textbf{\textcolor{mdred}{-7.511}}  &         0.000        &       \textbf{\textcolor{mdred}{-0.620}}    &       \textbf{\textcolor{mdred}{-0.356}}     \\
& \textbf{logC}      &      \textbf{\textcolor{mdred}{-0.4656}}  &        0.112     &    \textbf{\textcolor{mdred}{-4.145}}  &         0.000        &       \textbf{\textcolor{mdred}{-0.694}}    &       \textbf{\textcolor{mdred}{-0.237}}     \\

\bottomrule
\end{tabular}

%% file: tables/cv.tex
\begin{tabular}{ll | rrrr | rrr}
\toprule
\multirow{2}{*}{\textbf{Metric}} & \multirow{2}{*}{\textbf{Fold}} & \multicolumn{4}{c}{\textbf{Full}} & \multicolumn{3}{c}{\textbf{w/o Model Size}}\\
    &  & $\beta$ & $\gamma$ & $\theta$ & $R^2_{test}$ & $\beta$ & $\gamma$ & $R^2_{test}$ \\
\midrule
\multirow{5}{*}{\textbf{PS}} & 1  &  -0.3663  &  -0.6223  &  0.0130  &  0.7629  &  -0.3617  &  -0.6204  &  0.7653  \\
& 2  &  -0.3517  &  -0.5380  &  -0.0055  &  0.8969  &  -0.3527  &  -0.5387  &  0.8966  \\
& 3  &  -0.3607  &  -0.6232  &  0.0033  &  0.9282  &  -0.3600  &  -0.6222  &  0.9284  \\
& 4  &  -0.4000  &  -0.6023  &  0.0026  &  0.8338  &  -0.3992  &  -0.6019  &  0.8340  \\
& 5  &  -0.3775  &  -0.6438  &  -0.0167  &  0.8204  &  -0.3811  &  -0.6478  &  0.8234  \\
\midrule
\multirow{5}{*}{\textbf{LA}} & 1  &  -0.3713  &  -0.6599  &  0.0101  &  0.7629  &  -0.3677  &  -0.6584  &  0.7649  \\
& 2  &  -0.3627  &  -0.5386  &  -0.0127  &  0.8871  &  -0.3649  &  -0.5403  &  0.8864  \\
& 3  &  -0.3724  &  -0.6585  &  -0.0004  &  0.9056  &  -0.3725  &  -0.6586  &  0.9055  \\
& 4  &  -0.3933  &  -0.6364  &  -0.0032  &  0.8818  &  -0.3943  &  -0.6369  &  0.8816  \\
& 5  &  -0.3829  &  -0.6811  &  -0.0171  &  0.8073  &  -0.3866  &  -0.6852  &  0.8092  \\
\midrule
\multirow{5}{*}{\textbf{AR (accuracy)}} & 1  &  -0.5963  &  -0.7979  &  0.0176  &  0.7907  &  -0.5900  &  -0.7952  &  0.7942  \\
& 2  &  -0.5866  &  -0.6605  &  -0.0283  &  0.9060  &  -0.5916  &  -0.6643  &  0.9055  \\
& 3  &  -0.5762  &  -0.8585  &  0.0488  &  0.8115  &  -0.5670  &  -0.8437  &  0.8257  \\
& 4  &  -0.6115  &  -0.7897  &  -0.0189  &  0.9417  &  -0.6171  &  -0.7924  &  0.9418  \\
& 5  &  -0.5897  &  -0.7847  &  -0.1022  &  0.8923  &  -0.6117  &  -0.8093  &  0.9353  \\
\midrule
\multirow{5}{*}{\textbf{AR (mean rank)}} & 1  &  -0.5660  &  -0.4922  &  0.1891  &  0.7325  &  -0.4989  &  -0.4634  &  0.8427  \\
& 2  &  -0.4967  &  -0.5753  &  0.1073  &  0.7413  &  -0.4778  &  -0.5608  &  0.7146  \\
& 3  &  -0.5175  &  -0.4716  &  0.1340  &  0.8988  &  -0.4922  &  -0.4309  &  0.8876  \\
& 4  &  -0.5770  &  -0.4459  &  0.1659  &  0.8104  &  -0.5273  &  -0.4216  &  0.8597  \\
& 5  &  -0.4477  &  -0.4918  &  0.0377  &  0.6032  &  -0.4396  &  -0.4827  &  0.5866  \\
\bottomrule
\end{tabular}

%% file: tables/var.tex
\begin{tabular}{lcccccc|cccc}
\toprule
& \multicolumn{6}{c}{\gls{ih} emergence point $\log_{10}(U_{PT})$} & \multicolumn{4}{c}{Predictive law parameters} \\
\cmidrule(lr){2-7} \cmidrule(lr){8-11}
& (16,64) & (16,256) & (16,1024) & (16,2048) & (128,1024) & (512,1024) & $\alpha$ & $\beta$ & $\gamma$ & $R^2$ \\
\midrule
Mean & 4.308 & 3.828 & 3.507 & 3.458 & 2.987 & 2.910 & 13.540 & -0.470 & -0.589 & 0.968 \\
SD   & 0.060 & 0.064 & 0.029 & 0.008 & 0.023 & 0.034 & 0.401  & 0.026  & 0.048  & 0.005 \\
\bottomrule
\end{tabular}

%% file: tables/abab_example.tex
\begin{tabular}{l|cccccccccc}
\toprule
Word & \tone & \ttwo & \tthree & \tfour & \tfive & \tsix & \tone & \ttwo & \tfive & \tsix\\
\midrule
$R_U$ & 0 & 0 & 0 & 0 & 0 & 0 & 1 & 1 & 1 & 1 \\
$R_B$ & 0 & 0 & 0 & 0 & 0 & 0 & 1 & 0 & 1 & 0 \\
\bottomrule
\end{tabular}

%% file: tables/algo.tex
\begin{algorithm}[h]
\caption[Corpus generation constrained on \paba{A}{B} and \pbgivenaba{A}{B}]{Corpus generation constrained on \paba{A}{B} and \pbgivenaba{A}{B}}\label{ch5:alg:corpus}
\begin{algorithmic}[1]
\Statex \textit{Sample a token sequence} $\mathbf{s}$ \textit{from the distribution} $\mathcal{D}:\{w\mapsto \mathcal{D}_w\in\mathbb{R}^{|\mathcal{V}|}\mid w\in \mathcal{V}\}$,
\Statex \textit{constrained on} $\alpha=\paba{A}{B}$ and $\beta=\pbgivenaba{A}{B}$
\State \textbf{Input: }$\text{context\_size}$, $\mathcal{V}$, $\mathcal{D}\in\mathbb{R}^{|\mathcal{V}|\times|\mathcal{V}|}$, $\alpha$, $\beta$
\State $\mathbf{s} \gets [\space\space]$\Comment{init a token sequence}
\State $\mathcal{U}_t \gets \{\}$\Comment{init a unigram set at step $t$}
\State $\mathcal{B}_t \gets \{\}$\Comment{init a bigram continuation dict \{$w\mapsto \mathcal{B}_t(w)\mid w\in \mathcal{U}_t$\} at step $t$}
\State $m \gets \text{context\_size}//2$

\For{$t$ from 1 to $m$}\Comment{first half}
\State $w_{t} \sim\mathcal{D}_{w_{t-1}}(\mathcal{V}\backslash \mathcal{U}_{<t})$
\State $\text{add }w_t\text{ to }\mathbf{s}$
\State $\text{update }\mathcal{U}_t\text{ and }\mathcal{B}_t$
\EndFor
\For{$t$ from $m$ to context\_size}\Comment{second half}
\State is\_aba $\gets w_t \text{ in }\mathcal{U}_{t-1}$
\State make\_aba $\gets \text{random.random()}\le\alpha$\Comment{$\alpha$ represents \paba{A}{B}}
\State make\_abab $\gets \text{random.random()}\le\beta$\Comment{$\beta$ represents \pbgivenaba{A}{B}}
\If{is\_aba}
    \If{make\_abab}
        \State $w_t$ $\sim$ $\mathcal{D}_{w_{t-1}}(\cdot\mid \mathcal{B}_t(w_{t-1}))$
    \Else
        \If{make\_aba}
            \State $w_t\sim\mathcal{D}_{w_{t-1}}(\cdot\mid \mathcal{U}_{<t}\backslash \mathcal{B}_t(w_{t-1}))$
        \Else
            \State $w_t\sim \mathcal{D}(\cdot\mid \mathcal{V}\backslash\{\mathcal{U}\cup \mathcal{B}_t(w_{t-1})\})$
        \EndIf
    \EndIf

\Else
    \If{make\_aba}
        \State $w_t$ $\sim$ $\mathcal{D}_{w_{t-1}}(\cdot\mid \mathcal{U}_{<t})$
    \Else
        \State $w_{t} \sim\mathcal{D}_{w_{t-1}}(\cdot \mid \mathcal{V}\backslash \mathcal{U}_{<t})$
    \EndIf
\EndIf
\State $\text{add }w_t\text{ to }\mathbf{s}$
\State $\text{update }\mathcal{U}_t\text{ and }\mathcal{B}_t$
\EndFor
\end{algorithmic}
\end{algorithm}

%% file: tables/corpus_tok.tex
\begin{tabular}{ll@{\hspace{4pt}}llrrrrrrrrrr}
\toprule
 & & & & \multicolumn{10}{c}{\textbf{Sequence Length}} \\
\cmidrule(lr){5-14}
\textbf{Dataset} & \textbf{Tokenizer} & & & $4$ & $8$ & $16$ & $32$ & $64$ & $128$ & $256$ & $512$ & $1024$ & $2048$ \\
\midrule
\multirow{12}{*}{CC-100} & \multirow{4}{*}{LLaMA-3} & \multirow{2}{*}{$P_A$} & $\mu$ & 0.0052 & 0.0203 & 0.0516 & 0.1038 & 0.1709 & 0.2416 & 0.3111 & 0.3804 & 0.4503 & 0.5198 \\
 &  &  & $\sigma$ & 0.0363 & 0.0519 & 0.0580 & 0.0544 & 0.0451 & 0.0342 & 0.0257 & 0.0199 & 0.0158 & 0.0122 \\
\cmidrule(l){3-14}
 &  & \multirow{2}{*}{$P_B$} & $\mu$ & 0.0001 & 0.0023 & 0.0142 & 0.0395 & 0.0569 & 0.0660 & 0.0771 & 0.0943 & 0.1192 & 0.1514 \\
 &  &  & $\sigma$ & 0.0083 & 0.0343 & 0.0801 & 0.1052 & 0.0801 & 0.0525 & 0.0348 & 0.0239 & 0.0172 & 0.0126 \\
\cmidrule(l){2-14}
 & \multirow{4}{*}{GPT-2} & \multirow{2}{*}{$P_A$} & $\mu$ & 0.0058 & 0.0205 & 0.0519 & 0.1048 & 0.1728 & 0.2445 & 0.3153 & 0.3859 & 0.4565 & 0.5257 \\
 &  &  & $\sigma$ & 0.0386 & 0.0528 & 0.0590 & 0.0554 & 0.0458 & 0.0350 & 0.0265 & 0.0207 & 0.0163 & 0.0125 \\
\cmidrule(l){3-14}
 &  & \multirow{2}{*}{$P_B$} & $\mu$ & 0.0002 & 0.0024 & 0.0171 & 0.0492 & 0.0706 & 0.0820 & 0.0947 & 0.1133 & 0.1382 & 0.1691 \\
 &  &  & $\sigma$ & 0.0107 & 0.0357 & 0.0896 & 0.1206 & 0.0938 & 0.0644 & 0.0439 & 0.0304 & 0.0215 & 0.0155 \\
\cmidrule(l){2-14}
 & \multirow{4}{*}{T5} & \multirow{2}{*}{$P_A$} & $\mu$ & 0.0068 & 0.0237 & 0.0586 & 0.1167 & 0.1924 & 0.2718 & 0.3478 & 0.4209 & 0.4926 & 0.5616 \\
 &  &  & $\sigma$ & 0.0415 & 0.0563 & 0.0625 & 0.0583 & 0.0480 & 0.0361 & 0.0266 & 0.0200 & 0.0154 & 0.0119 \\
\cmidrule(l){3-14}
 &  & \multirow{2}{*}{$P_B$} & $\mu$ & 0.0002 & 0.0036 & 0.0247 & 0.0669 & 0.0946 & 0.1105 & 0.1246 & 0.1428 & 0.1678 & 0.1999 \\
 &  &  & $\sigma$ & 0.0103 & 0.0429 & 0.1042 & 0.1295 & 0.0945 & 0.0624 & 0.0407 & 0.0276 & 0.0194 & 0.0141 \\
\midrule
\multirow{12}{*}{FineWeb} & \multirow{4}{*}{LLaMA-3} & \multirow{2}{*}{$P_A$} & $\mu$ & 0.0054 & 0.0227 & 0.0591 & 0.1200 & 0.2018 & 0.2918 & 0.3778 & 0.4548 & 0.5216 & 0.5797 \\
 &  &  & $\sigma$ & 0.0371 & 0.0572 & 0.0713 & 0.0769 & 0.0749 & 0.0693 & 0.0643 & 0.0602 & 0.0553 & 0.0488 \\
\cmidrule(l){3-14}
 &  & \multirow{2}{*}{$P_B$} & $\mu$ & 0.0001 & 0.0046 & 0.0309 & 0.0866 & 0.1374 & 0.1739 & 0.2045 & 0.2326 & 0.2585 & 0.2817 \\
 &  &  & $\sigma$ & 0.0084 & 0.0485 & 0.1197 & 0.1634 & 0.1477 & 0.1255 & 0.1081 & 0.0939 & 0.0823 & 0.0709 \\
\cmidrule(l){2-14}
 & \multirow{4}{*}{GPT-2} & \multirow{2}{*}{$P_A$} & $\mu$ & 0.0062 & 0.0234 & 0.0607 & 0.1253 & 0.2121 & 0.3051 & 0.3920 & 0.4686 & 0.5344 & 0.5914 \\
 &  &  & $\sigma$ & 0.0394 & 0.0579 & 0.0716 & 0.0761 & 0.0729 & 0.0673 & 0.0626 & 0.0585 & 0.0538 & 0.0479 \\
\cmidrule(l){3-14}
 &  & \multirow{2}{*}{$P_B$} & $\mu$ & 0.0001 & 0.0044 & 0.0333 & 0.0981 & 0.1574 & 0.2023 & 0.2378 & 0.2676 & 0.2929 & 0.3143 \\
 &  &  & $\sigma$ & 0.0086 & 0.0474 & 0.1247 & 0.1708 & 0.1491 & 0.1226 & 0.1028 & 0.0873 & 0.0739 & 0.0631 \\
\cmidrule(l){2-14}
 & \multirow{4}{*}{T5} & \multirow{2}{*}{$P_A$} & $\mu$ & 0.0079 & 0.0279 & 0.0689 & 0.1381 & 0.2301 & 0.3274 & 0.4173 & 0.4957 & 0.5633 & 0.6218 \\
 &  &  & $\sigma$ & 0.0450 & 0.0636 & 0.0760 & 0.0784 & 0.0731 & 0.0654 & 0.0583 & 0.0533 & 0.0484 & 0.0427 \\
\cmidrule(l){3-14}
 &  & \multirow{2}{*}{$P_B$} & $\mu$ & 0.0003 & 0.0059 & 0.0396 & 0.1077 & 0.1655 & 0.2085 & 0.2441 & 0.2753 & 0.3031 & 0.3279 \\
 &  &  & $\sigma$ & 0.0122 & 0.0546 & 0.1329 & 0.1699 & 0.1447 & 0.1191 & 0.1001 & 0.0858 & 0.0737 & 0.0639 \\
\midrule
\multirow{12}{*}{Dolma} & \multirow{4}{*}{LLaMA-3} & \multirow{2}{*}{$P_A$} & $\mu$ & 0.0052 & 0.0222 & 0.0587 & 0.1221 & 0.2087 & 0.3044 & 0.3945 & 0.4739 & 0.5423 & 0.6016 \\
 &  &  & $\sigma$ & 0.0361 & 0.0547 & 0.0659 & 0.0689 & 0.0663 & 0.0622 & 0.0585 & 0.0556 & 0.0520 & 0.0477 \\
\cmidrule(l){3-14}
 &  & \multirow{2}{*}{$P_B$} & $\mu$ & 0.0001 & 0.0033 & 0.0242 & 0.0742 & 0.1230 & 0.1627 & 0.1987 & 0.2331 & 0.2645 & 0.2927 \\
 &  &  & $\sigma$ & 0.0066 & 0.0408 & 0.1043 & 0.1439 & 0.1252 & 0.1046 & 0.0908 & 0.0812 & 0.0733 & 0.0658 \\
\cmidrule(l){2-14}
 & \multirow{4}{*}{GPT-2} & \multirow{2}{*}{$P_A$} & $\mu$ & 0.0059 & 0.0224 & 0.0590 & 0.1253 & 0.2161 & 0.3138 & 0.4047 & 0.4841 & 0.5522 & 0.6105 \\
 &  &  & $\sigma$ & 0.0382 & 0.0550 & 0.0662 & 0.0686 & 0.0648 & 0.0602 & 0.0563 & 0.0532 & 0.0500 & 0.0459 \\
\cmidrule(l){3-14}
 &  & \multirow{2}{*}{$P_B$} & $\mu$ & 0.0001 & 0.0032 & 0.0268 & 0.0855 & 0.1436 & 0.1908 & 0.2307 & 0.2652 & 0.2955 & 0.3214 \\
 &  &  & $\sigma$ & 0.0067 & 0.0400 & 0.1111 & 0.1544 & 0.1300 & 0.1047 & 0.0877 & 0.0759 & 0.0672 & 0.0603 \\
\cmidrule(l){2-14}
 & \multirow{4}{*}{T5} & \multirow{2}{*}{$P_A$} & $\mu$ & 0.0069 & 0.0254 & 0.0653 & 0.1356 & 0.2315 & 0.3332 & 0.4265 & 0.5073 & 0.5766 & 0.6361 \\
 &  &  & $\sigma$ & 0.0416 & 0.0585 & 0.0687 & 0.0694 & 0.0642 & 0.0577 & 0.0526 & 0.0486 & 0.0450 & 0.0410 \\
\cmidrule(l){3-14}
 &  & \multirow{2}{*}{$P_B$} & $\mu$ & 0.0002 & 0.0043 & 0.0329 & 0.0958 & 0.1517 & 0.1974 & 0.2377 & 0.2744 & 0.3074 & 0.3363 \\
 &  &  & $\sigma$ & 0.0091 & 0.0467 & 0.1205 & 0.1551 & 0.1264 & 0.1005 & 0.0841 & 0.0732 & 0.0651 & 0.0587 \\
\midrule
\multirow{12}{*}{The Pile} & \multirow{4}{*}{LLaMA-3} & \multirow{2}{*}{$P_A$} & $\mu$ & 0.0083 & 0.0349 & 0.0872 & 0.1669 & 0.2660 & 0.3696 & 0.4656 & 0.5486 & 0.6183 & 0.6760 \\
 &  &  & $\sigma$ & 0.0462 & 0.0752 & 0.1047 & 0.1242 & 0.1330 & 0.1323 & 0.1270 & 0.1185 & 0.1085 & 0.0973 \\
\cmidrule(l){3-14}
 &  & \multirow{2}{*}{$P_B$} & $\mu$ & 0.0004 & 0.0147 & 0.0721 & 0.1724 & 0.2629 & 0.3270 & 0.3758 & 0.4164 & 0.4514 & 0.4816 \\
 &  &  & $\sigma$ & 0.0145 & 0.0872 & 0.1801 & 0.2268 & 0.2164 & 0.2029 & 0.1956 & 0.1888 & 0.1806 & 0.1693 \\
\cmidrule(l){2-14}
 & \multirow{4}{*}{GPT-2} & \multirow{2}{*}{$P_A$} & $\mu$ & 0.0807 & 0.1172 & 0.1702 & 0.2476 & 0.3429 & 0.4399 & 0.5281 & 0.6038 & 0.6668 & 0.7183 \\
 &  &  & $\sigma$ & 0.2036 & 0.2202 & 0.2167 & 0.2071 & 0.1940 & 0.1767 & 0.1586 & 0.1413 & 0.1248 & 0.1082 \\
\cmidrule(l){3-14}
 &  & \multirow{2}{*}{$P_B$} & $\mu$ & 0.0538 & 0.0961 & 0.1707 & 0.2672 & 0.3518 & 0.4150 & 0.4631 & 0.5019 & 0.5350 & 0.5629 \\
 &  &  & $\sigma$ & 0.1775 & 0.2471 & 0.2971 & 0.3036 & 0.2778 & 0.2521 & 0.2333 & 0.2183 & 0.2045 & 0.1894 \\
\cmidrule(l){2-14}
 & \multirow{4}{*}{T5} & \multirow{2}{*}{$P_A$} & $\mu$ & 0.0250 & 0.0621 & 0.1236 & 0.2122 & 0.3202 & 0.4297 & 0.5277 & 0.6097 & 0.6773 & 0.7317 \\
 &  &  & $\sigma$ & 0.0942 & 0.1160 & 0.1312 & 0.1402 & 0.1415 & 0.1359 & 0.1275 & 0.1174 & 0.1059 & 0.0925 \\
\cmidrule(l){3-14}
 &  & \multirow{2}{*}{$P_B$} & $\mu$ & 0.0063 & 0.0227 & 0.0838 & 0.1893 & 0.2867 & 0.3618 & 0.4203 & 0.4675 & 0.5079 & 0.5418 \\
 &  &  & $\sigma$ & 0.0626 & 0.1152 & 0.1904 & 0.2243 & 0.2105 & 0.1958 & 0.1886 & 0.1831 & 0.1773 & 0.1677 \\
\bottomrule
\end{tabular}